\documentclass[numbers,webpdf]{ima-authoring-template}%

\newcommand{\E}{\mathbb{E}}
\graphicspath{{Fig/}}

\theoremstyle{thmstyletwo}%
\newtheorem{theorem}{Theorem}
\newtheorem{remark}{Remark}%
\newtheorem{definition}{Definition}
\usepackage{amssymb}

\usepackage{amsmath}
\newtheorem{corollary}{Corollary}

\newtheorem{lemma}{Lemma}
\newtheorem{proof}{Proof}

\usepackage{makecell,colortbl,booktabs,multirow}

\usepackage{enumitem}

\usepackage{natbib}

\usepackage{color, xcolor}

\usepackage{algorithm}
\usepackage{algpseudocode}
\usepackage[utf8]{inputenc} 
\usepackage[T1]{fontenc}    
\usepackage{hyperref}       
\usepackage{url}            
\usepackage{booktabs}       
\usepackage{amsfonts}       
\usepackage{nicefrac}       
\usepackage{microtype}      

\numberwithin{equation}{section}

\begin{document}

\DOI{DOI HERE}
\copyrightyear{2025}
\vol{00}
\pubyear{2025}
\access{Advance Access Publication Date: Day Month Year}
\appnotes{Paper}
\copyrightstatement{Published by Oxford University Press on behalf of the Institute of Mathematics and its Applications. All rights reserved.}


\title{Batched Bandits with Heavy-Tailed Rewards}




\author{
    Yunwen Guo
    \address{\orgdiv{School of Mathematical Sciences}, \orgname{Fudan University}, \orgaddress{\street{220 Handan Rd.}, \postcode{200433}, \state{Shanghai}, \country{China}}}
}
\author{Yunlu Shu
    \address{\orgdiv{Shanghai Center for Mathematical Sciences}, \orgname{Fudan University}, \orgaddress{\street{220 Handan Rd.}, \postcode{200433}, \state{Shanghai}, \country{China}}}
}
\author{Gongyi Zhuo
    \address{\orgdiv{School of Mathematical Sciences}, \orgname{Fudan University}, \orgaddress{\street{220 Handan Rd.}, \postcode{200433}, \state{Shanghai}, \country{China}}}
}
\author{Tianyu Wang*
    \address{\orgdiv{Shanghai Center for Mathematical Sciences}, \orgname{Fudan University}, \orgaddress{\street{220 Handan Rd.}, \postcode{200433}, \state{Shanghai}, \country{China}}}
}

\corresp[*]{Corresponding author: \href{email:wangtianyu@fudan.edu.cn}{wangtianyu@fudan.edu.cn}}

\received{Date}{0}{Year}
\revised{Date}{0}{Year}
\accepted{Date}{0}{Year}


\abstract{The batched multi-armed bandit (MAB) problem, where rewards are collected in batches, is pivotal in applications like clinical trials. While prior work assumes light-tailed reward distributions, real-world scenarios often exhibit heavy-tailed outcomes. This paper addresses this gap by introducing robust batched bandit algorithms for heavy-tailed rewards in both multi-arm and Lipschitz settings. We uncover somewhat surprising phenomena for such problems -- heavier tails require \textit{fewer} batches to achieve near-optimal regret in the instance-independent setting, as well as the Lipschitz setting. In sharp contrast, in the instance-dependent setting, the number of batches required to achieve near-optimal regret does not dependent on the tail heaviness. }



\maketitle


\section{Introduction} 

The multi-armed bandit (MAB) problem, a cornerstone of sequential decision-making, has found widespread applications in clinical trials, recommendation systems, and adaptive resource allocation. Among its many variants, the batched bandit problem has garnered significant attention in recent years, particularly due to its relevance in clinical trial design \cite{pocock1977group,mahajan2010adaptive,pallmann2018adaptive,perchet2016batched}. Unlike classical bandit settings where decisions are updated after every observation, batched bandits restrict the algorithm to a limited number of policy updates, reflecting real-world constraints where data arrives in batches rather than a continuous stream.



Indeed, in trials for new drugs or therapies, patients are typically enrolled in cohorts, with safety and efficacy data analyzed between batches to adjust dosages, treatment arms, or even terminate ineffective interventions early. While prior work on batched bandits has provided valuable insights into optimal exploration-exploitation trade-offs under batch constraints \cite{gao2019batched,feng2022lipschitz}, most existing studies rely on a critical yet unrealistic assumption: that rewards follow light-tailed distributions (e.g., sub-Gaussian or bounded variables).

However, real-world clinical outcomes often exhibit heavy-tailed behavior \cite{lee2003analyzing,cantoni2006robust}. Patient responses to treatments can display high variability, with rare but extreme outcomes (e.g., severe side effects or unusually strong recoveries) that defy standard tail assumptions. Similar challenges arise in financial decision-making, disaster response planning, and other high-stakes domains where outliers carry significant weight. Ignoring heavy-tailed rewards can lead to overly optimistic algorithms that fail in practice, as they underestimate the risk of extreme events. 

Despite its practical importance, the batched bandit problem with heavy-tailed rewards remains understudied. Although seminal works \citep{thompson1933likelihood,perchet2016batched,gao2019batched} have ingeniously laid the foundation for (stochastic) bandits and batched bandits, existing approaches either (1) focus on stochastic rewards with sub-Gaussian tails or (2) study heavy-tailed bandits only in the fully sequential setting. Bridging this gap is crucial for developing robust clinical trial protocols --- and more generally, for deploying bandit algorithms in safety-critical applications where outcomes are batched and rewards are volatile.

To this end, we study the batched bandit problem under heavy-tailed noise and develop a robust batched bandit algorithm. Specifically, we consider the following setting:  
The agent has a set of arms $\mathcal{X}$ to choose from. Selecting an arm $x \in \mathcal{X}$ yields an $i.i.d.$ reward drawn from a probability distribution $\nu_x$ with mean $\mathbb{E}[\nu_x] = \mu_x$. Additionally, we assume there exist constants $\varepsilon \in (0,1]$ and $v \in (0,\infty)$ such that $\mathbb{E}\left[|\nu_x - \mu_x|^{1+\varepsilon}\right] \le v$ for all $x \in \mathcal{X}$.  

We examine two key scenarios: 1. Finite-arm setting: $\mathcal{X} = \{1, 2, \dots, K\}$ for some $K < \infty$. 2. Lipschitz setting: $\mathcal{X}$ is a compact metric space, allowing for structural assumption (Lipschitzness) on the mean reward function.

Our results are summarized in Tables \ref{tab:summary} and \ref{tab:lipschitz}. Surprisingly, when the noise is more heavy-tailed, less batches are required to achieve a near-optimal instance-independent regret. 


\begin{table}[h!] 
    \caption{Summary of our results for the finite-arm setting. 
    }
    \centering 
    \begin{tabular*}{\textwidth}{@{\extracolsep\fill}c|cc@{\extracolsep\fill}} 
    \toprule 
     & Instance-independent results & Instance-dependent results \\ \hline 
    \makecell{Regret\\upper bound} & $ v^\frac{1}{1+\varepsilon} \log K \left( K \log (KT) \right)^{\frac{\varepsilon}{1+\varepsilon}} T^{\frac{1}{1+\varepsilon-\varepsilon\left( \frac{\varepsilon}{1+\varepsilon} \right)^{M-1}}} $ & $ \frac{v^\frac{1}{\varepsilon} (\log K) K }{(\min_{i \neq \star} \triangle_i)^{\frac{1}{\varepsilon}}} \log (TK) T^\frac{1}{M} $ \\ \hline 
    \makecell{Min. \# batches\\to achieve\\near-opt. regret} & $  \frac{\log \log T}{ \log \left( \frac{1+\varepsilon} {\varepsilon} \right) } $ & $  \log T  $ \\ \hline 
    \makecell{Regret\\lower bound} & $K^\frac{\varepsilon}{1+\varepsilon} T^{\frac{1}{1+\varepsilon-\varepsilon\left( \frac{\varepsilon}{1+\varepsilon} \right)^{M-1}}}$ & $\left( \frac{1}{\min_{i \neq \star} \triangle_i}  \right)^{\frac{1}{\varepsilon}} K T^\frac{1}{M} $ \\ \hline 
    \makecell{Min. \# batches\\for near-opt.\\regret to\\be achievable} & $  \frac{ \log \log T }{ \log \left( \frac{1+\varepsilon} {\varepsilon} \right) } $ & $\log T$ \\ 
    \bottomrule 
    \end{tabular*}  
    \begin{tablenotes}%
    \item In the table, $T$ is the total time horizon, $M$ is number of batches, and $ \varepsilon \in (0,1] $ is the heavy-tail parameter --- the $ (1+\varepsilon) $-th moment of the rewards are bounded. 
    All cells display values in terms of order; constants are neglected. 
    \end{tablenotes}
    \label{tab:summary} 
\end{table}

\begin{table}[!ht] 
    \caption{Summary of our results for the Lipschitz setting.  } 
    \centering
    \begin{tabular*}{\textwidth}{@{\extracolsep\fill}c|c@{\extracolsep\fill}} 
    \toprule 
    \makecell{Regret\\upper bound} & $ v^\frac{1}{\varepsilon} \log \log \frac{T}{\log T} (\log T)^{\frac{1}{d_z+1+\frac{1}{\varepsilon}}} T^{\frac{d_z+\frac{1}{\varepsilon}}{d_z+1+\frac{1}{\varepsilon}}} $ \hspace{8em}  
    \\ \hline 
    \makecell{Min. \# batches\\to achieve\\near-opt. regret} & $\frac{\log \log \frac{T}{\log T}}{\log (d+1+\frac{1}{\varepsilon}) -\log (d+1-d_z)}$ \hspace{8em}
    \\ \hline 
    \makecell{Regret\\lower bound} & $ T^{\frac{1-\frac{1}{d_z+1+\frac{1}{\varepsilon}}}{1-\left( \frac{1}{d_z+1+\frac{1}{\varepsilon}}\right)^M}} $ \hspace{8em}
    \\ \hline 
    \makecell{Min. \# batches\\for near-opt.\\regret to\\be achievable} & $  \frac{ \log \log T }{ \log \left( d_z+1+\frac{1}{\varepsilon} \right) } $  \hspace{8em}
    \\ 
    \bottomrule 
    \end{tabular*} 
    \begin{tablenotes}%
    \item In the table, $T$ is the total time horizon, $M$ is number of batches, and $ \varepsilon \in (0,1] $ is the heavy-tail parameter --- the $ (1+\varepsilon) $-th moment of the rewards are bounded.
    \end{tablenotes}
    \label{tab:lipschitz} 
\end{table} 

In summary, our main contributions are: 
\begin{itemize}[leftmargin=*]
    \item \textbf{Robust Batched Bandit Algorithms for Heavy-Tailed Rewards:} We develop batched bandit algorithms capable of handling heavy-tailed reward distributions, in both the finite-arm and the Lipschitz settings. This addresses a critical gap in the original motivation behind batched bandits --- real-world applications like clinical trials often exhibit heavy-tailed outcomes, yet prior work largely assumes light-tailed rewards. Our algorithms provide a principled solution to these underexplored problems.
    \item \textbf{Surprising Findings on Tail Heaviness and Communication Patterns.}
Our analysis reveals key relationship between reward tail heaviness and optimal communication patterns in batched bandit learning, with surprising implications for both instance-independent and instance-dependent settings.
\begin{itemize}[leftmargin=*]
    \item \textbf{Instance-independent Setting: A Counterintuitive Phenomenon.}
We discover a counterintuitive result: as rewards become more heavy-tailed (i.e., as $\varepsilon$ decreases), \textbf{fewer communication batches are needed to achieve near-optimal regret.} This appears paradoxical at first glance -- since heavy-tailed rewards yield noisier, less informative samples, one might expect more frequent communication (and thus more adaptation opportunities) would be necessary. Yet, our analysis reveals the opposite. Increasing the number of batches fails to mitigate the inherent informational deficit caused by heavy tails; instead, the optimal response is to communicate less frequently. 
Remarkably, this phenomenon extends to Lipschitz bandits, where the interplay between reward sample information and the arm space's metric structure creates an even more nuanced relationship. 

\item 
\textbf{Instance-Dependent Setting: A Sharp Contrast.}
In sharp contrast, the optimal communication pattern in the instance-dependent setting exhibits no dependence on the heavy-tail parameter $\varepsilon$. 
\end{itemize} 
\end{itemize}

\begin{remark}
    The phenomenon in the instance-independent setting -- as well as in the Lipschitz setting -- can be understood as follows: increasing the number of batches reduces the sample size per batch, which in turn leads to larger estimation errors under heavy-tailed distributions. To mitigate the risk of discarding the optimal arm, the optimal strategy for heavier tails is to use fewer communication rounds. What makes the phenomenon more intriguing, however, is that tail heaviness does not affect the communication pattern in the instance-dependent case. This discrepancy likely arises because the communication grids for the two cases are scattered very differently; see Figure \ref{fig:comm-grids} for an illustration. 
    
\end{remark}

Our results reveal a striking divergence between instance-independent and instance-dependent settings, demonstrating how tail behavior differentially impacts learning:
    \textbf{Instance-Independent Setting}: The tail parameter $\varepsilon$ fundamentally governs the exploration-communication trade-off. Heavier tails (smaller $\varepsilon$) lead to sparser optimal communication schedules. 
    \textbf{Instance-Dependent Setting}: Local problem geometry (suboptimality gaps) determines the optimal communication pattern. While tail heaviness affects the overall regret bounds, it does not influence the batch allocation strategy.

\subsection{Related works}

The stochastic Multi-Armed Bandit (MAB) problem has a rich theoretical history dating back several decades \citep{thompson1933likelihood,robbins1952some}. Seminal works established its theoretical foundations \citep{gittins1979bandit,lai1985asymptotically}, while later research developed practical algorithms with strong theoretical guarantees \citep{auer2002finite,auer2002nonstochastic}. Over this research history, several canonical algorithms have emerged, including Thompson sampling \citep{thompson1933likelihood,agrawal2012analysis}, UCB method \citep{lai1985asymptotically,auer2002finite}, and exponential weight \citep{auer2002nonstochastic,arora2012multiplicative}, with extra methods available in recent texts \citep{MAL-024,MAL-068,lattimore2020bandit}. 

Motivated by the growing demand for large-scaled data \cite{berry1985bandit,cesa2013online}, the batched feedback setting has emerged as a critical research area, particularly for medical scenarios requiring periodic data analysis. Perchet et al. \cite{perchet2016batched} launched the study of batched bandit problems, with \cite{gao2019batched} later resolving key challenges in batched multi-armed bandits. Recent years have seen significant advances in batched bandit learning \citep{jun2016top,agarwal2017learning,tao2019collaborative,han2020sequential,karpov2020collaborative,esfandiari2021regret,ruan2021linear,li2022gaussian,agarwal2022batched,jin2021almost,jin2021double}.

Recent years have witnessed growing interest in bandit problems with heavy-tailed distributions. Vakili et al. \cite{vakili2013deterministic} investigated both sample mean and truncated mean estimators for heavy-tailed rewards. Building upon this work, Bubeck et al. \cite{bubeck2013bandits} conducted a more comprehensive study of various mean estimators under weak moment conditions. Their analysis established upper and lower bounds for the bandit problem, advancing our theoretical understanding of heavy-tailed bandits.
The advances in heavy-tailed bandits span multiple directions. \cite{lugosi2021robust}  examines robust mean estimation in reward analysis. \cite{yu2018pure} and \cite{shao2018almost} address pure exploration and linear bandits respectively. Regret minimization is advanced by \cite{wei2020minimax} through minimax policies and by \cite{agrawal2021regret} via statistical-computational trade-offs.

The Lipschitz bandit problem was originally introduced as "continuum-armed bandits" \citep{agrawal1995continuum}, where the arm set is a compact interval. Subsequent work extended this setting to bandits with Lipschitz (or H\"{o}lder) continuous reward functions. \cite{kleinberg2005nearly} established a $\Omega(T^{2/3})$ lower bound for this problem and proposed an algorithm that matches it. When additional conditions beyond Lipschitzness are imposed, a regret rate of $\widetilde{\mathcal{O}}(T^{1/2})$ becomes achievable \citep{auer2007improved, cope2009regret}. For general metric spaces with doubling properties, algorithms such as the Zooming bandit \citep{kleinberg2008multi} and Hierarchical Optimistic Optimization (HOO) \citep{bubeck2011x} have been developed. Subsequently, many variants of Lipschitz bandit problems have been extensively studied \citep{bubeck2011lipschitz,10.1145/3412815.3416885,lu2019optimal,slivkins2011contextual,krishnamurthy2020contextual,majzoubi2020efficient,10960481}. Among many, the most related ones are the BLiN algorithm \cite{feng2022lipschitz} coauthored by the last author, and Lipschitz bandit with heavy-tailed noise \citep{lu2019optimal}. 
Our work generalizes the ACE sequence -- a key contribution of \cite{feng2022lipschitz} -- to a tail-heaviness-aware version that unifies metric entropy for bandits in metric spaces, tail heaviness for robust mean estimators, and communication patterns for batched feedback.

Despite the extensive body of existing bandit literature, our work uncovers findings that have not been reported elsewhere: 
the influence of tail-heaviness affects the grid spacing very differently in the instance-dependent and instance-independent settings. Specifically, heavier tails suppress frequent communication in the instance-independent case, yet they do not alter the communication pattern in the instance-dependent case. 
Furthermore, this effect observed in the instance-independent case carries over to the Lipschitz bandit setting. 

\subsection{Organization}

The remainder of this paper is structured as follows.
Section \ref{section:pre} introduces fundamental concepts and preliminaries. Section \ref{section:alg} presents our batched bandit algorithms for both multi-armed and Lipschitz settings. Section \ref{section:upper} establishes theoretical guarantees for these algorithms, including regret upper bounds.
Meanwhile, Section \ref{section:upper} provides the theoretical discussions along with illustrative examples.
The corresponding proofs of the main results in Section \ref{section:upper} are provided in Section \ref{app:upper}. Section \ref{section:lower} derives the lower bound for the batched bandit problem, with supporting proofs presented in Section \ref{app:lower}. Numerical experiments are presented in Section \ref{experiments}.

\section{Preliminaries} 
\label{section:pre}

In stochastic bandit learning, an agent makes sequential decisions and observes noisy rewards associated with each choice. The agent faces a dual challenge: it must simultaneously learn the expected reward structure while trying to maximize cumulative rewards. The agent's performance is measured through regret -- the cumulative difference between the optimal rewards and those actually obtained. For a policy $\pi$ over $T$ steps, the regret is: 
\begin{align*} 
R_T(\pi) := \E \left[ \sum_{t=1}^T \mu_{\star} - \mu_{x_t} \right] ,
\end{align*} 
where $x_t \in \mathcal{X}$ is the policy's choice at step $t$, $\mu_x$ is the expected reward at $x$ (for any $x \in \mathcal{X}$), and $\star$ is the optimal arm. In this paper, we focus on problems with heavy-tailed rewards, where pulling an arm $x \in \mathcal{X}$ yields an independent reward drawn from a distribution $\nu_{x}$ such that $ \E \nu_{x} = \mu_x $ and $ \E | \nu_x - \mu_x |^{1 + \varepsilon} \le v $ for some constants $\varepsilon \in (0,1]$ and $v \in (0,\infty)$. 

The batched setting introduces additional constraints. Reward observations are collected in batches and only become available at predetermined or adaptively chosen communication time points. 
Specifically, the time horizon is partitioned via a strictly increasing sequence of communication time points: 
$\mathcal{T} := \{t_j\}_{j=0}^M$, where $ 0 = t_0 < t_1 < \cdots < t_M = T $ {and} $ M \ll T$. 
Under this protocol: For any $t \in (t_{j-1}, t_j]$, the reward sample at $t$ becomes observable only at $t_j$. 
As a result, the policy's action at time $t$, denoted $x_t$, is $\mathcal{F}_{t_{j-1}}$-measurable, where $\mathcal{F}_{t_{j-1}}$ is the $\sigma$-algebra generated by all historical information up to $t_{j-1}$.

\subsection{Robust mean estimators}
We first recall properties of mean estimators for an unknown distribution $X$. Let $X_1,\cdots, X_n$ be $i.i.d.$ copies of  $X$. Let $X$ have finite $(1+\varepsilon)$-th moment: $\mathbb{E}\left[ | X-\mu |^{1+\varepsilon} \right] \le v < \infty $, where $ \mu = \E X $ is the mean of $X$. 
For such random variables, we can design mean estimators that exhibit concentration properties generalizing those of sub-Gaussian distributions. This can be summarized as follows.

\begin{lemma}[Lemma 2 in \cite{bubeck2013bandits}] 
    \label{heavy-tail:property}
    Consider distribution $X$ with finite mean $\mu$ and
    finite $1+\varepsilon$ moments 
    $\mathbb{E}\left[ | X-\mu |^{1+\varepsilon} \right] \le v$ for some parameters $\varepsilon \in (0,1]$ and $v \in (0, \infty)$.
    Let $X_1,\cdots,X_n$ be i.i.d. random variables following $X$.
    For any $\delta \in (0,1)$ and $n$ there exists robust mean estimator $\hat{\mu} = \hat{\mu}(n,\delta)$, such that, with probability at least $1-\delta$, 
        $ \left| \hat{\mu} - \mu \right| \le v^{1 /(1+\varepsilon)}\left(\frac{c \log (1 / \delta)}{n}\right)^{\varepsilon /(1+\varepsilon)}, $
    for some constant $c$.  
\end{lemma}

A concrete mean estimator that satisfies Lemma \ref{heavy-tail:property} is the median-of-means estimator \cite{ALON1999137,Lugosi2019-ap}, whose definition can be found in Definition \ref{def:mom}. 


\begin{definition}[Median of means] 
    \label{def:mom}
    Let $X_1, \ldots, X_n$ be i.i.d. random variables with mean $\mathbb{E} X=\mu$ and centered $(1+\varepsilon)$-th moment $\mathbb{E}|X-\mu|^{1+\varepsilon} \le v$ for $\varepsilon \in(0,1]$. Choose $k=\left\lfloor 8 \log \left(e^{1 / 8} / \delta\right) \wedge n / 2\right\rfloor$, here $\delta \in(0,1)$ is small. Divide $n$ random variables into groups of size $N=\lfloor n / k\rfloor$ and
    \begin{align*} 
        \hat{\mu}_1=\frac{1}{N} \sum_{t=1}^N X_t, \quad \hat{\mu}_2=\frac{1}{N} \sum_{t=N+1}^{2 N} X_t, \quad \ldots, \quad \hat{\mu}_k=\frac{1}{N} \sum_{t=(k-1) N+1}^{k N} X_t 
    \end{align*} 
    are $k$ empirical mean estimates. Median of median $\widehat{\mu}_M (n,\delta)$ are defined as median of these empirical means. 
\end{definition}

Building on prior works \cite{ALON1999137,Lugosi2019-ap}, the median-of-means estimator is known to be robust and satisfies the concentration inequalities in Lemma \ref{heavy-tail:property}. Other estimators -- such as the truncated mean \cite{Bickel1965-wr} and Catoni's $M$-estimator \cite{10.1214/11-AIHP454} -- also achieve these bounds. 

In this work, for any arm $x \in \mathcal{X}$ and sample size $s$, we define $\hat{\mu}_{x,s} := \hat{\mu}_x(s, \delta)$ as a robust mean estimator constructed from $s$ $i.i.d.$ reward samples, ensuring compliance with Lemma \ref{heavy-tail:property}.  

\subsection{Zooming dimension}

For the Lipschitz bandit problem, the arm set is a compact doubling metric space $(\mathcal{X},d_{\mathcal{X}})$, and without loss of generality, the expected reward function is a 1-Lipschitz function with respect to the metric $d_{\mathcal{X}}$:   $|\mu_x - \mu_y| \le d_{\mathcal{X}} (x, y)$ for any $x, y \in \mathcal{X} $. 
The optimal reward is $\mu_\star = \max_{x \in A} \mu_x $. For any arm $x$, the optimality gap is $\triangle_x = \mu_\star - \mu_x $. 

\begin{definition}
    For a problem instance with arm set $\mathcal{X}$ and expected reward $\mu$, we let $S(r)$ denote $r$-level set $S(r)=\left\{x \in \mathcal{X}: \triangle_x \leq r\right\}$. We define $N_r=\mathcal{N}\left(S(16 r), r\right)$ as the $r$-packing number to pack $S(16 r)$ with $r$-balls. The zooming dimension is then defined as
    \begin{align*}
        d_z:=\min \left\{d \ge 0: \exists a>0, N_r \leq a r^{-d}, \forall 0<r<1\right\} .
    \end{align*}
    Moreover, the zooming constant $C_z$ is $ C_z=\min \left\{a>0: N_r \leq a r^{-d_z}, \forall 0<r<1\right\} $. 
        
\end{definition} 

\begin{remark}
    In this work, we focus on the metric space $([0,1]^d, \| \cdot \|_\infty)$, where metric balls are cubes. 
    From an algorithmic perspective, this choice does not limit generality, since all procedures presented for $([0,1]^d, \| \cdot \|_\infty)$ can be easily generalized to other compact doubling metric spaces. In Lipschitz setting, we use the notation $\mu(x) : = \mu_x$. 
\end{remark}

\section{The Algorithms} 
\label{section:alg}



A defining feature of batched bandit algorithms is their communication pattern -- the timing of data collection. Our findings reveal that distributions with heavier tails favor fewer batches, implying that the optimal communication schedule depends on the heavy-tail parameter $\varepsilon$. To formalize this, we need define communication time points explicitly tied to $\varepsilon$.  

Once the communication pattern is set, the algorithm proceeds as follows:  1. \textbf{Uniform Exploration:} Within each batch, the agent treats all arms equally.  
2. \textbf{Adaptive Elimination}: After collecting rewards, arms (or regions) deemed suboptimal with high confidence are discarded.  

\subsection{Finite-arm Setting: explicitly specifying the communication time points}

Consider a $T$-step batched bandit game with $M$ allowed communication rounds. The grid of communication time points $\mathcal{T} = \{t_0, t_1, t_2, \dots, t_M = T\}$ are defined as follows. 

\textbf{The Instance-independent case.} Let $l = T^{\frac{1}{1+\varepsilon-\varepsilon\left( \frac{\varepsilon}{1+\varepsilon} \right)^{M-1}}}$ be the base scaling factor, where $\varepsilon$ is the heavy-tail parameter. The communication time points are then determined by: 
\begin{align}
    t_1 = l, \quad 
    t_m = l \cdot (t_{m-1})^{\frac{\varepsilon}{1+\varepsilon}} \quad \text{for} \quad m = 2,\dots,M, \label{eq:grid-indep}
\end{align}
and $t_0 = 0$ is the boundary convention. With the above definition, we have $t_m = l^{1+\varepsilon-\varepsilon \left( \frac{\varepsilon}{1+\varepsilon} \right)^{m-1}}$ and $t_M = T$. We define this grid of time points by $\mathcal{T}^1$.

Naturally, these grid shall be integers, so we perform a rounding operation $\left \lfloor t_m \right \rfloor$ when the algorithm is carried out. Since this rounding operation does not change regret order, we use $t_m$ to simplify calculations. 

\textbf{The Instance-dependent case.} Define $l^{\prime} = T^{\frac{1}{M}}$, and let 
\begin{align}
    t_0^{\prime}=0, \quad t_1^{\prime}=l^{\prime}, \quad t_{m}^{\prime} = l^{\prime} \cdot t_{m-1}^{\prime}, \quad m = 2,\cdots,M, \label{eq:grid-dep}
\end{align}
be the communication time points. With this definition, we have $ t_m' = T^{\frac{m}{M}} $. We define this grid of time points by $\mathcal{T}^2$.

\begin{figure}[!h]
  \centering
  \includegraphics[width=0.92\linewidth]{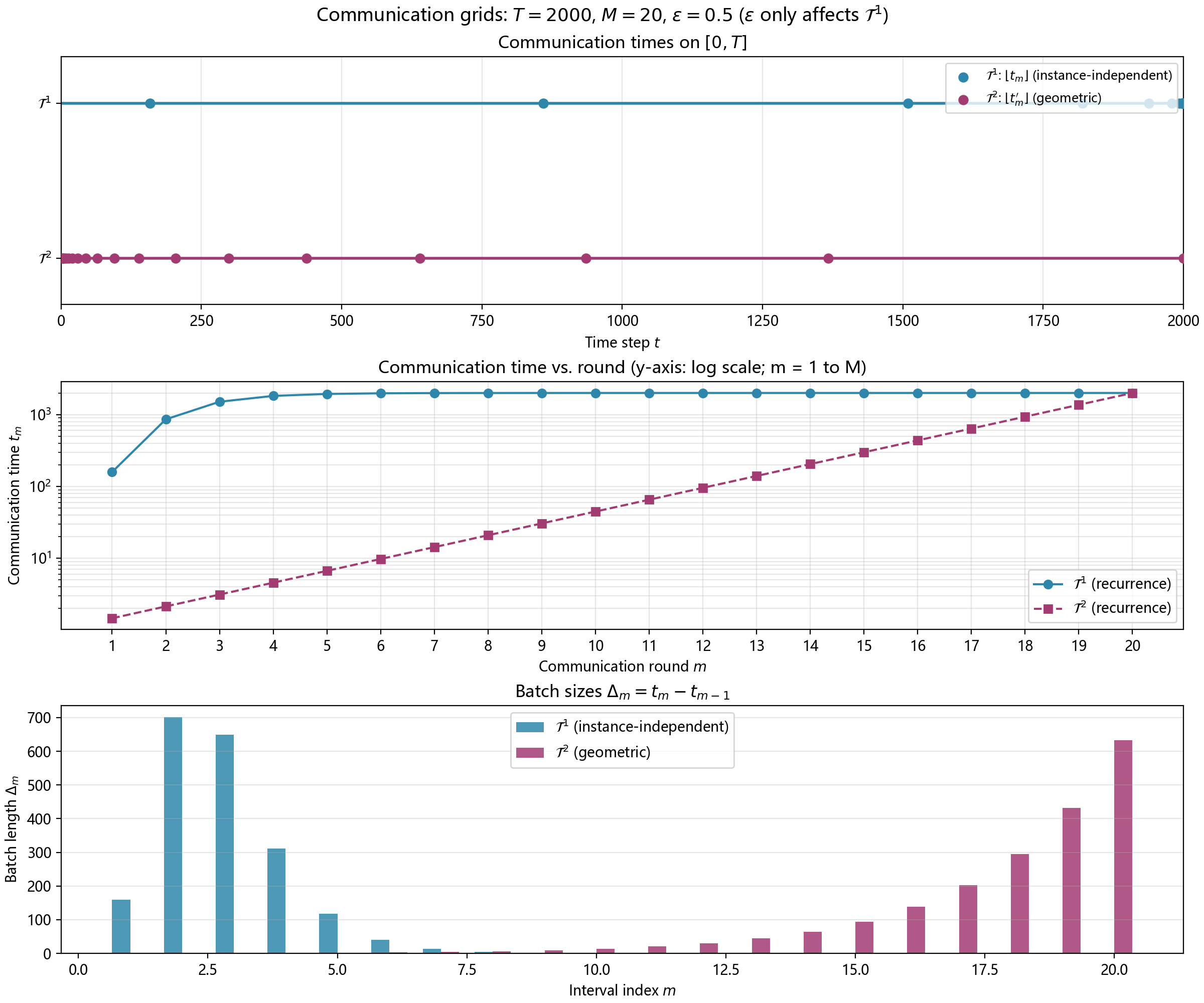}
  \caption{
    Comparison of communication-time grids $\mathcal{T}^1$
    (instance-independent) and $\mathcal{T}^2$ (geometric) for horizon $T$,
    $M$ communication rounds, and parameter $\epsilon$ (only $\mathcal{T}^1$
    depends on $\epsilon$). Top: communication times along $[0,T]$ (floor
    rounding). Middle: $t_m$ vs.\ round $m$ with a logarithmic vertical axis
    ($m=1,\ldots,M$). Bottom: batch lengths $\Delta_m=t_m-t_{m-1}$.}
  \label{fig:comm-grids}
\end{figure}



The communication time grids for the instance-independent strategy $\mathcal{T}^1$ and the geometric strategy $\mathcal{T}^2$ are visualized in FIG \ref{fig:comm-grids}. The top panel illustrates the distribution of communication time points over the horizon $T=2000$. It is evident that $\mathcal{T}^1$ (blue dots) maintains a relatively uniform distribution of communication events across the timeline. In contrast, $\mathcal{T}^2$ (purple dots) exhibits a ``front-loaded'' density, where communication points are clustered closely during the initial stages and become increasingly sparse as time progresses.

This behavior is quantified in the middle panel, which plots the communication time $t_m$ against the communication round $m$ on a logarithmic scale. The geometric strategy $\mathcal{T}^2$ follows a distinct exponential growth curve, indicating that the intervals between communications expand significantly in later rounds. Conversely, $\mathcal{T}^1$ shows a steadier, near-linear progression on this scale. Consequently, as shown in the bottom panel, the batch length $\Delta_m$ (the interval between consecutive communications) for $\mathcal{T}^2$ starts small and increases drastically in later rounds, whereas $\mathcal{T}^1$ maintains a more consistent batch size distribution.

\subsection{Lipschitz Setting: implicit communication time points via the narrowing principle} 
\label{sec:lip-algo}
For Lipschitz bandits, the communication pattern is coupled with the narrowing strategy \cite{feng2022lipschitz,10239433,10960481}, refining the approach as follows:  

\textbf{Uniform Play:} At the start of batch $m$, the arm space is partitioned into a set of \textit{active cubes} $\mathcal{A}_m = \{C_{m,1}, C_{m,2}, \dots, C_{m, |\mathcal{A}_m|}\}$, where each cube $C \in \mathcal{A}_m$ has edge length $r_m$ (construction details follow in the next step).  

During batch $m$, each cube $C \in \mathcal{A}_m$ is sampled  
\begin{align*}
    n_m \triangleq \frac{16 \log T}{r_m^2}
\end{align*}
times, where $T$ is the time horizon. Specifically, for each $C$, the center of $C$ (denoted $ \texttt{center} (C)$) is played for $n_m$ times, and their rewards $\{y_{C,1}, \dots, y_{C,n_m}\}$ are recorded at the batch’s end.  

\textbf{Arm Elimination and Partitioning:} After batch $m$, the mean reward $\widehat{\mu}_m(C)$ of each cube $C$ is estimated as:  
\begin{align*}
    \widehat{\mu}_m(C) = \frac{1}{n_m} \sum_{i=1}^{n_m} y_{C,i}. 
\end{align*} 
A cube $C$ is eliminated if  
\begin{align*}
    \widehat{\mu}_m^{\max} - \widehat{\mu}_m(C) \geq 4 r_m, 
\end{align*} 
where $\widehat{\mu}_m^{\max} = \max_{C' \in \mathcal{A}_m} \widehat{\mu}_m(C')$. 
Each surviving cube is partitioned into $(\frac{r_m}{r_{m+1}})^d$ subcubes with diameter (edge-length) $r_{m+1}$, where $\{r_m\}_{m=1}^M$ is a predefined decreasing sequence (with appropriate rounding to maintain integer partition ratios). These new subcubes constitute the active set $\mathcal{A}_{m+1}$ for the next batch iteration.

\textbf{Diameter Sequence Specification ($\varepsilon$-ACE Sequence). } 
In the Lipschitz bandit setting, the communication pattern is governed by the diameter sequence 
$\{ r_m \}$, which generalizes the ACE Sequence introduced in \citep{feng2022lipschitz} to a tail‑shape‑aware formulation. This edge length sequence serves as a bridge among metric entropy, tail heaviness, and communication grid spacing. 
For heavy-tailed rewards with parameter $\varepsilon$, this edge length sequence  must incorporate the tail behavior and is formally defined as follows:

\begin{definition}[$\varepsilon$-ACE Sequence]
The sequence $\{r_m\}$ is given by:
$r_m = 2^{-\sum_{i=1}^m c_i}$,
where the coefficients $\{c_i\}$ satisfy:
$c_1 = \frac{d_z + \frac{1}{\varepsilon}}{(d + 1 + \frac{1}{\varepsilon})(d_z + 1 + \frac{1}{\varepsilon})} \log (\frac{T}{\log T})$,
and subsequent terms follow the recurrence relation:
$c_{i+1} = \eta c_i$ for $i \geq 1$, with $\eta = \frac{d + 1 - d_z}{d + 1 + \frac{1}{\varepsilon}}$.
\label{def:lips-rm}
\end{definition}


Once the diameter (edge-length) sequence $\{r_m\}$ is specified , the algorithm automatically determines: the number of samples each cube: $n_m$, which subsequently determines the adaptive communication schedule $\{t_m\}$.  
The communication time points emerge dynamically as a consequence of the geometric partitioning, with batch transitions occurring upon completing all required pulls within the current active set $\mathcal{A}_m$.

The communication pattern serves as the foundation for adapting bandit algorithms to heavy-tailed rewards. Once this pattern is established -- whether explicitly defined or implicitly determined through the diameter sequence -- we can modify existing algorithms to accommodate heavy-tailed distributions: For finite-armed setting: We adapt the BaSE algorithm \cite{perchet2016batched,gao2019batched}. For Lipschitz setting: We extend the BLiN framework \cite{feng2022lipschitz,10239433}. 
The resulting heavy-tailed algorithms are presented formally in Algorithms \ref{alg:base} and \ref{alg:algorithm-lips}. 
To accommodate more complex scenarios where different arms have distinct parameters $\varepsilon_i$, we make slight modifications to Algorithms \ref{alg:base} in the Appendix.






\begin{algorithm}[ht]
\caption{Batched Successive Elimination for Heavy-Tailed Bandits (BaSE-H)}\label{alg:base}
\begin{algorithmic}[1]
\State \textbf{Input:} Arms $\mathcal{I} = \{1, 2, \dots, K\}$; Time horizon $T$; Number of batches $M$; Grid of time points $\mathcal{T} = \{t_1, \dots, t_M\}$; Heavy-tailed parameters $\varepsilon$ and $v$.
\State Initialize active arm set $A \gets \mathcal{I}$; $t_0 \gets 0$.

\For{$m \gets 1$ \textbf{to} $M - 1$}
    \State \textbf{Batch Execution:} During $[t_{m-1} + 1, t_m]$, pull arms in $A$ circularly s.t.\ $|\tau_i(t_m) - \tau_j(t_m)| \leq 1$ for all $\{i,j\} \subset A$. Define $\tau(t_m) \triangleq \min_{i\in A} \tau_i(t_m) $.
    \State Compute $\hat{\mu}_i(\tau(t_m))$ for $i = 1, \dots, K$. Let $\hat{\mu}_{\max}(\tau(t_m)) \gets \max_{j \in A} \hat{\mu}_j(\tau(t_m))$.
    
    \For{each arm $i \in A$}
        \If{$\hat{\mu}_{\max}(\tau(t_m)) - \hat{\mu}_i(\tau(t_m)) \geq 2^{\frac{1+2\epsilon}{1+\epsilon}} v^{\frac{1}{1+\epsilon}} \left(\frac{c \log(TK)}{\tau(t_m)}\right)^{\frac{\epsilon}{1+\epsilon}}$}
            \State $A \gets A \setminus \{i\}$ \qquad \qquad \qquad \qquad \qquad \qquad \qquad \qquad /* {Eliminate suboptimal arm} */
        \EndIf
    \EndFor
\EndFor

\State \textbf{Clean-up Phase:}
\State Identify $i_0 \gets \mathrm{argmax}_{j \in A} \hat{\mu}_j(\tau_j(t_{M-1}))$.
\For{$t \gets t_{M-1} + 1$ \textbf{to} $T$}
    \State Pull arm $i_0$ exclusively.
\EndFor

\State \textbf{Output:} Regret $\pi$.
\end{algorithmic}
\end{algorithm}

\begin{algorithm}[ht]
\caption{Batched Lipschitz Narrowing for Heavy-Tailed Bandit}\label{alg:algorithm-lips}
\begin{algorithmic}[1]
\State \textbf{Input:} Arm $[0,1]^d$; Time horizon $T$; Number of batches $M$; Heavy-tailed parameters $\varepsilon$ and $v$; Sequence $\{r_m\}_{m=1}^{M+1}$. 
\State Initialize active arm set $A \gets [0,1]^d$; $t_0 \gets 0$; Equally partition $A$ to $r_1^d$ subcubes and define $A_1$ as the collection of these subcubes.
\State Compute $n_m = 3c v^{\frac{1}{\varepsilon}} \log T \cdot r_m^{-\frac{1+\varepsilon}{\varepsilon}} $ for $m=1,\cdots, M+1$.

\For{$m \gets 1$ \textbf{to} $M - 1$}
    \State \textbf{Batch Execution:} During $m$-th batch, pull each cube $ C \in A_m$ $n_m$ times. Collect the rewards of all pulls up to $t_m$. 
    \State Compute $\hat{\mu}_C(n_m)$ for each cube $C \in A_m$. Let $\hat{\mu}_{\max}(n_m) \gets \max_{C \in A_m} \hat{\mu}_C(n_m)$. 
    \State Initialize $A_{m}^+ = A_m$.
    
    \For{each arm $C \in A_m$}
        \If{$\hat{\mu}_{\max}(n_m) - \hat{\mu}_C(n_m) \geq 4 r_m$}
            \State $A_m^+ \gets A_m^+ \setminus \{C\}$ \qquad \qquad \qquad \qquad \qquad \qquad \qquad /* {Eliminate suboptimal arm} */
        \EndIf
    \EndFor
    \State Now $A_m^+$ is the set of cubes not eliminated after $m$-th batch. Compute $t_{m+1} = t_m + (r_m/r_{m+1})^d \cdot |A_m^+| \cdot n_{m+1}$.
    \If{$t_{m+1} \ge T$} 
        \State \textbf{break}
    \EndIf
    \State Equally partition each cube in $A_m^+$ into $(r_m/r_{m+1})^d$ subcubes. Define $A_{m+1}$ as collection of these subcubes.
\EndFor

\State \textbf{Clean-up Phase:} Play the remaining arms exclusively until all $T$ steps used up.

\State \textbf{Output:} Regret $\pi$.
\end{algorithmic}
\end{algorithm}

\section{Performance Guarantees}
\label{section:upper}


Algorithms \ref{alg:base} and \ref{alg:algorithm-lips} satisfy the properties stated in the following theorems. 

\begin{theorem}
    Consider a $K$-armed batched bandit problem where the batched number is $M$. 
    Suppose the distributions of all arms have $(1+\varepsilon)$-th moment bounded by the same constant $v < \infty$. 
    Let $\pi^1$ be the policy equipped with the communication time points $\mathcal{T}^1$ defined in \eqref{eq:grid-indep} and $\pi^2$ be the policy equipped with communication time points $\mathcal{T}^2$ defined in \eqref{eq:grid-dep}. Both policies apply Algorithm \ref{alg:base}. For $\max _{i \in[K]} \Delta_i \le  K^{\frac{\varepsilon}{1+\varepsilon}}$, and $T \in \mathbb{N}_+$, we have
    \begin{align}
        & \mathbb{E}\left[R_T\left(\pi^1\right)\right] \le \kappa v^\frac{1}{1+\varepsilon} \log K \left( K \log (TK) \right)^{\frac{\varepsilon}{1+\varepsilon}} T^{\frac{1}{1+\varepsilon-\varepsilon\left( \frac{\varepsilon}{1+\varepsilon} \right)^{M-1}}}, \label{without gap} \\
        & \mathbb{E}\left[R_T\left(\pi^2\right)\right] \le \kappa  \left( \frac{1}{\min \triangle_i}\right)^{\frac{1}{\varepsilon}} v^\frac{1}{\varepsilon} \log(K) K \log (TK) T^\frac{1}{M}, \label{with gap} 
    \end{align}
where $\kappa >0$ is a numerical constant independent of $\varepsilon, v, K, M$ and $T$.  
    \label{thm:upper}
\end{theorem}

\begin{corollary}
    Instate the conditions in Theorem \ref{thm:upper}, the gap-independent regret $\mathbb{E}\left[R_T\left(\pi^1\right)\right]$ for Algorithm \ref{alg:base} has upper bound of order $\mathcal{O}\left( \log (T) ^{\frac{\varepsilon}{1+\varepsilon}} T^{\frac{1}{1+\varepsilon-\varepsilon\left( \frac{\varepsilon}{1+\varepsilon} \right)^{T-1}}} \right)$ with only $\Omega \left( \log^{-1} \left( \frac{1+\varepsilon}{\varepsilon} \right) \log \log T \right)$ number of batches.
    \label{cor:upper-M}
\end{corollary}

\begin{theorem}
    Consider a $T$-step Lipschitz bandit problem where the reward distributions have $(1 + \varepsilon)$-th moment bounded by $v< \infty$. 
    Let $\pi$ be the policy equipped with Algorithm \ref{alg:algorithm-lips} and sequence $\{r_m\}$ in Definition \ref{def:lips-rm}. For $T > 0$, we have 
    \begin{align*}
        \mathbb{E}[R_T(\pi)] \le
        \kappa v^\frac{1}{\varepsilon} C_z \log \log \frac{T}{\log T} (\log T)^{\frac{1}{d_z+1+\frac{1}{\varepsilon}}} T^{\frac{d_z+\frac{1}{\varepsilon}}{d_z+1+\frac{1}{\varepsilon}}},
    \end{align*}
    where $\kappa > 0$ is a numerical constant independent of any parameter and $C_z$ is the zooming constant. Furthermore, this regret needs only $\mathcal{O} \left(\frac{\log \log \frac{T}{\log T}}{\log (d+1+\frac{1}{\varepsilon}) -\log (d+1-d_z)}\right)$ number of batches.
    \label{thm:upper-lips}
\end{theorem} 

As the moment parameter $\varepsilon$ decreases, we obtain less information about the heavy-tailed distribution. This robustness comes at the cost of increased regret.
Proofs of the theorems and corollary rely on properties of robust mean estimators and design of algorithms. The proof for the finite-arm setting is presented in the next section, while the analysis for the Lipschitz case is deferred to the Appendix.

\subsection{Findings and An Illustrative Example}

These theorems reveal a consistent phenomenon: under both the multi-armed bandit (instance-independent case) and Lipschitz bandit settings, heavier-tailed rewards require fewer communication batches to achieve near‑optimal regret. This can be understood intuitively as follows.

Increasing the number of batches typically forces each batch to be allocated fewer samples, which in turn leads to larger estimation errors due to the slow convergence rate under heavy-tailed distributions. To mitigate this risk, the optimal strategy is to control the number of communication rounds, allowing each batch to accumulate a sufficient number of samples to achieve reliable estimation, sacrificing adaptivity for robustness.


We will illustrate this phenomenon with two Pareto distributions as a toy example, with $K=2$ arms and total horizon $T=10^6$. Recall that the communication time points for batch 
$M$ are as defined in Section \ref{section:alg}. The only difference is the heavy‑tail parameter $\epsilon$, which determines the $(1+\epsilon)$-th moment bound and the convergence rate of robust mean estimators.

\textbf{Case A: Moderately heavy tail $( \varepsilon=0.9)$}

Consider a random variable $X$ $\sim$ Pareto $(\alpha=1.91)$ with support $[1,\infty)$. 
Its expectation is $\mu$ and its $(1+\varepsilon)$-th moment is finite for any $\varepsilon \le 0.9$. In particular, its $1.9$-th moment exists:
$$
\mathbb{E}\bigl[|X-\mu|^{1.9}\bigr] \le v < \infty,
$$
where $v$ is a constant.
From Lemma \ref{heavy-tail:property}, the robust estimator $\hat{\mu}$ of $X$ satisfies the following error bound: With probability $1-\delta$,
$$
|\hat\mu-\mu| \le 
v^{1/1.9}\left(\frac{c\log(1/\delta)}{n}\right)^{0.9/1.9}.
$$
Hence the error decays as $n^{-0.474}$.
To ensure that the error $|\hat\mu-\mu|$ is bounded by  $\theta$, we require 
$$
n \ge \left(\frac{v^{1/1.9}(c\log(1/\delta))^{0.9/1.9}}{\theta}\right)^{1.9/0.9} \propto \theta^{-2.111}.
$$

From Corollary \ref{cor:upper-M} of the paper, for $\epsilon=0.9$ and $T=10^6$, the optimal number of batches $M_{\min}$ is
$$
M_{\min} \approx \frac{1}{0.746}\times 1.7918 \approx 2.40 \quad\Longrightarrow\quad \text{about }2\text{ batches}.
$$

\textbf{Case B: Extremely heavy tails ($\epsilon = 0.1$)}

Consider a random variable $X$ $\sim$ Pareto $(\alpha=1.11)$ with support $[1,\infty)$. 
Its expectation is $\mu$ and its $1.1$-th centered moment exists:
$$
\mathbb{E}\bigl[|X-\mu|^{1.1}\bigr] \le v < \infty.
$$
where $v$ is a constant.
Following the same calculation procedure, the optimal number of batches for $\epsilon=0.1$ is
$$
M_{\min} \approx \frac{1}{2.398}\times 1.7918 \approx 0.747 \quad\Longrightarrow\quad \text{about }1\text{ batch}.
$$
We compare two cases with the following Table \ref{tab:comparison}. The table shows that although heavier tails degrade the estimation accuracy, they also reduce the benefit of frequent communication. The optimal strategy shifts toward using fewer batches, each with enough samples to obtain a trustworthy estimate. This counterintuitive phenomenon is mathematically captured by the dependence of the batch schedule on $\epsilon$ and is reflected in the regret bounds of the paper.

\begin{table}[ht]
\centering      
              
\begin{tabular}{|l|c|c|}
\hline
\textbf{Aspect} & \textbf{Case A (moderate tails)} & \textbf{Case B (extreme tails)} \\ \hline
Pareto shape $\alpha$ & $1.91$ & $1.11$ \\ \hline
Heavy-tail parameter $\epsilon$ & $0.9$ & $0.1$ \\ \hline
Moment condition & $\mathbb{E}[|X-\mu|^{1.9}]<\infty$ & $\mathbb{E}[|X-\mu|^{1.1}]<\infty$ \\ \hline
Error convergence rate & $n^{-0.474}$ & $n^{-0.0909}$ \\ \hline
Sample size for error $\theta$ & $\propto \theta^{-2.111}$ & $\propto \theta^{-11}$ \\ \hline
Optimal batches $M_{\min}$ & $\approx 2$ & $\approx 1$ \\ \hline
\end{tabular}
\caption{Comparison of two Pareto distributions with different tail heaviness under the finite-arm batched bandit setting ($K=2$, $T=10^6$).}
\label{tab:comparison}

 \end{table} 
\section{Upper bound analysis for the finite-arm setting}
\label{app:upper}

\subsection{Preliminary Analysis}

We first state that with high probability the best arm is not eliminated and the other arms will not be chosen too many times.

Let $B$ be the event that the best arm $\star$ is not eliminated throughout the time horizon $T$. For $i \ne \star$, let $\tau_i^{\ast} \triangleq 2^{\frac{2+3\varepsilon}{\varepsilon}} v^{\frac{1}{\varepsilon}} c \log (TK) \left( \frac{1}{\triangle_i} \right)^{\frac{1+\varepsilon}{\varepsilon}}$ and define the following cut-off:
\begin{align*}
    m_i=\min \left\{j \in[M]: \operatorname{arm} i \text { has been pulled at least } \tau_i^{\ast} \text { times before time } t_j \in \mathcal{T}\right\},
\end{align*}
supplementally define $m_i=M$ if the minimum does not exist and $m_\star=M$.
Let $A_i$ be the event that arm $i$ is eliminated before time $t_{m_i}$, and we set $A_i$ occurs if $m_i=M$.
Define $E$ as the event that all these events happen: $E = B \cap \left( \cap_{i=1}^K A_i \right)$, which is the final event that holds with high probability.

\begin{lemma}
    The event $E$ happens with probability at least $1-\frac{4}{TK}$ in \textbf{Algorithm} \ref{alg:base}.
    \label{lem:upper}
\end{lemma}

When the event $E$ does not occur, the expected regret is at most 
\begin{align*}
    \mathbb{E} \left[R_T\left(\pi\right) \mathbb{I}(E^c) \right] \le T \max_{i \in [K]} \triangle_i \cdot \mathbb{P}(E^c) = \mathcal{O}(1).
\end{align*}
Thus we can assume that $E$ holds. Proof of Lemma \ref{lem:upper} is provided in Appendix \ref{app:lem}. We now proceed directly to prove the main theorem.

\subsection{Proof of Theorem \ref{thm:upper}}

\begin{proof}

We cut the proof into three parts. Thanks to Lemma \ref{lem:upper}, we only need to consider regret under event $E$.

\subsubsection*{Part I}
In this part, we analyze regret under communication grid $\mathcal{T}^1$ defined in Eq. (\ref{eq:grid-indep}).
For policy $\pi^1$, let $\mathcal{I}_0 \subseteq \mathcal{I}$ be the subset of arms which are eliminated at the end of the first batch, $\mathcal{I}_1 \subseteq \mathcal{I}$ be the subset of remaining arms which are eliminated before the last batch, and $\mathcal{I}_2=\mathcal{I}-\mathcal{I}_0-\mathcal{I}_1$ be the subset of arms which remain in the last batch. 

It is clear that the total regret incurred by arms in $\mathcal{I}_0$ is at most $t_1 \cdot \max _{i \in[K]} \Delta_i \le K^{\frac{\varepsilon}{1+\varepsilon}} T^{\frac{1}{1+\varepsilon-\varepsilon\left( \frac{\varepsilon}{1+\varepsilon} \right)^{M-1}}} $ because $t_1 = l = T^{\frac{1}{1+\varepsilon-\varepsilon\left( \frac{\varepsilon}{1+\varepsilon} \right)^{M-1}}}$ in $\mathcal{T}^1$. It remains to deal with the sets $\mathcal{I}_1$ and $\mathcal{I}_2$ separately.

For $\mathcal{I}_1$, consider arm $i \in \mathcal{I}_1$, let $n_i$ be the number of arms which are eliminated before arm $i$. Observe that the fraction of pullings of arm $i$ is at least $\frac{1}{K}$, at most $\frac{1}{K-n_i}$ before arm $i$ is eliminated. 
For precise calculation, define $\tau_i$ the number of pullings of $i$ when arm $i$ is eliminated.
Because arm $i$ is eliminated before $t_{m_i}$, there is $\tau_i \le \frac{t_{m_i}}{K-n_i}$. Under event $E$, $\tau_i^{\ast} > \tau_i$.
Moreover, by the definition of $\tau_i^{\ast}$, we must have
\begin{align}
    \tau_i^{\ast} > \tau_i \geq \frac{t_{m_{i-1}}}{K} -1 \Longrightarrow 
    2^{\frac{2+3\varepsilon}{\varepsilon}} v^{\frac{1}{\varepsilon}} c \log (TK) \left( \frac{1}{\triangle_i} \right)^{\frac{1+\varepsilon}{\varepsilon}} \ge \frac{t_{m_{i-1}}}{K}.
    \label{eq:upper-tri_i}
\end{align} 

Hence, the total regret incurred by pulling an arm $i \in \mathcal{I}_1$ is at most
\begin{align*}
    R_T(\pi^1)_i & \le \triangle_i \cdot \tau_i \le \triangle_i \frac{t_{m_i}}{K-n_i} \\
    & \le 2^\frac{2+3\varepsilon}{1+\varepsilon} v^{\frac{1}{1+\varepsilon}} \left( c \log (TK) \cdot K \right)^{\frac{\varepsilon}{1+\varepsilon}} \cdot \frac{1}{(t_{m_{i-1}})^\frac{\varepsilon}{1+\varepsilon}} \cdot \frac{t_{m_i}}{K-n_i} \\
    & \le 2^\frac{2+3\varepsilon}{1+\varepsilon} v^{\frac{1}{1+\varepsilon}} \left( c \log (TK) K \right)^{\frac{\varepsilon}{1+\varepsilon}} l \frac{1}{K-n_i} .
\end{align*}
Here we use the inequality obtained from Eq. (\ref{eq:upper-tri_i}) and the definition in $\mathcal{T}^1$ : $t_m = l \cdot (t_{m-1})^\frac{\varepsilon}{1+\varepsilon}$.
Notice that $\{ n_i\}_{i=1}^K$ is a permutation of $\left[K\right]$, $\sum \frac{1}{K-n_i} \sim \sum \frac{1}{t}$. The total regret incurred by $\mathcal{I}_1$ is at most
\begin{align*}
    R_T(\pi^1) \mathcal{I}_1 
    \le \sum_{i \in \mathcal{I}_1} R_T(\pi^1)_i
    \le 2^\frac{2+3\varepsilon}{1+\varepsilon} v^{\frac{1}{1+\varepsilon}} \left( c \log (TK) K \right)^{\frac{\varepsilon}{1+\varepsilon}} l \log K.
\end{align*}

For $\mathcal{I}_2$, consider arm $i \in \mathcal{I}_2$, also define $\tau_i$ as the number of pullings of $i$ when arm $i$ is eliminated. Since $m_i = M$ for $i \in \mathcal{I}_2$, similar to the analysis of $i \in \mathcal{I}_1$, we have
\begin{align*}
    \tau_i^{\ast} > \tau_i(t_{M-1}) \ge \frac{t_{M-1}}{K} -1 \Longrightarrow 
    2^{\frac{2+3\varepsilon}{\varepsilon}} v^{\frac{1}{\varepsilon}} c \log (TK) \left( \frac{1}{\triangle_i} \right)^{\frac{1+\varepsilon}{\varepsilon}} \ge \frac{t_{M-1}}{K} = \left(\frac{T}{l}\right)^{\frac{1+\varepsilon}{\varepsilon}} \frac{1}{K}.
\end{align*}
Hence, the total regret incurred by pulling an arm $i \in \mathcal{I}_2$ is at most
\begin{align*}
    R_T(\pi^1)_i & \le \triangle_i \cdot \tau_i \\
    & \le \tau_i \cdot 2^\frac{2+3\varepsilon}{1+\varepsilon} v^{\frac{1}{1+\varepsilon}} \left( c \log (TK) \cdot K \right)^{\frac{\varepsilon}{1+\varepsilon}} \cdot \frac{l}{T}.
\end{align*}
Since $\sum_{\mathcal{I}_2} \tau_i \le T $ , the total regret incurred by pulling arms in $\mathcal{I}_2$ is at most
\begin{align*}
    R_T(\pi^1) \mathcal{I}_2 & \le 2^\frac{2+3\varepsilon}{1+\varepsilon} v^{\frac{1}{1+\varepsilon}} \left( c \log (TK) K \right)^{\frac{\varepsilon}{1+\varepsilon}} l.
\end{align*}

Add different regret part together, we have
\begin{align*}
    \mathbb{E}\left[R_T\left(\pi^1\right)\right] & \le \mathbb{E} \left[R_T\left(\pi^1 \right) \mathbb{I}(E^c)\right] \mathbb{P}(E^c) + \mathbb{E} \left[R_T\left(\pi^1 \right) \mathbb{I}(E)\right] \mathbb{P}(E) \\
    & \le O(1) + R_T(\pi^1) \mathcal{I}_0 + R_T(\pi^1) \mathcal{I}_1 +R_T(\pi^1) \mathcal{I}_2 \\
    & \le O(1) + K^{\frac{\varepsilon}{1+\varepsilon}} T^{\frac{1}{1+\varepsilon-\varepsilon\left( \frac{\varepsilon}{1+\varepsilon} \right)^{M-1}}} + 2^\frac{2+3\varepsilon}{1+\varepsilon} v^{\frac{1}{1+\varepsilon}} \left( c \log (TK) K \right)^{\frac{\varepsilon}{1+\varepsilon}} l \log K \\
    & \quad + 2^\frac{2+3\varepsilon}{1+\varepsilon} v^{\frac{1}{1+\varepsilon}} \left( c \log (TK) K \right)^{\frac{\varepsilon}{1+\varepsilon}} l \\
    & \le C v^{\frac{1}{1+\varepsilon}} \left( \log (TK) K \right)^{\frac{\varepsilon}{1+\varepsilon}} \log K T^{\frac{1}{1+\varepsilon-\varepsilon\left( \frac{\varepsilon}{1+\varepsilon} \right)^{M-1}}}, 
\end{align*}
where $C > 0 $ is a numerical constant, which does not depend on any parameter.

\subsubsection*{Part II}
For policy $\pi^2$, our communication grid $\{t_0^{\prime},\cdots,t_M^{\prime}  \}$ follows $\mathcal{T}^2$ in Eq. (\ref{eq:grid-dep}). We follow the notation in Part I. The underlying methodology also aligns with the strategy developed in Part I.

The total regret incurred by arms in $\mathcal{I}_0$ is at most $t_1^{\prime} \cdot \max _{i \in[K]} \Delta_i \le K^{\frac{\varepsilon}{1+\varepsilon}} T^{\frac{1}{M}} $.

For $\mathcal{I}_1$, consider arm $i \in \mathcal{I}_1$, the total regret incurred by arm $i \in \mathcal{I}_1$ is at most
\begin{align*}
    R_T(\pi^2)_i & \le \triangle_i \frac{t_{m_i}^{\prime}}{K-n_i} \le \triangle_i^{-\frac{1}{\varepsilon}} \cdot \triangle_i^{\frac{1+\varepsilon}{\varepsilon}} \frac{t_{m_i}^{\prime}}{K-n_i} \\
    & \le \triangle_i^{-\frac{1}{\varepsilon}} 2^\frac{2+3\varepsilon}{\varepsilon} v^{\frac{1}{\varepsilon}} c \log (TK) \cdot \frac{K}{t_{m_{i-1}}^{\prime}} \cdot \frac{t_{m_i}^{\prime}}{K-n_i} \\
    & \le \triangle_i^{-\frac{1}{\varepsilon}} 2^\frac{2+3\varepsilon}{\varepsilon} v^{\frac{1}{\varepsilon}} c \log (TK) \cdot l^{\prime} \cdot \frac{K}{K-n_i} .
\end{align*}
The total regret incurred by $\mathcal{I}_1$ is at most
\begin{align*}
    R_T(\pi^2) \mathcal{I}_1 & \le \sum_{i \in \mathcal{I}_1} \triangle_i^{-\frac{1}{\varepsilon}} 2^\frac{2+3\varepsilon}{\varepsilon} v^{\frac{1}{\varepsilon}} c \log (TK) K T^\frac{1}{M}.
\end{align*}

For $\mathcal{I}_2$, the total regret incurred by pulling an arm $i \in \mathcal{I}_2$ is at most
\begin{align*}
    R_T(\pi^2)_i & \le \triangle_i \cdot \tau_i \le \triangle_i (\tau_i^{\ast}+T-t_{M-1}^{\prime}) \\
    & \le \triangle_i \tau_i^{\ast} + \triangle_i^{-\frac{1}{\varepsilon}} \triangle_i^{\frac{1+\varepsilon}{\varepsilon}} (T-t_{M-1}^{\prime}) \\
    & \le \triangle_i^{-\frac{1}{\varepsilon}} 2^{\frac{2+3\varepsilon}{\varepsilon}} v^{\frac{1}{\varepsilon}} c \log (TK) + \triangle_i^{-\frac{1}{\varepsilon}} 2^\frac{2+3\varepsilon}{\varepsilon} v^{\frac{1}{\varepsilon}} c \log (TK) \cdot \frac{K}{t_{M-1}^{\prime}} (T-t_{M-1}^{\prime}) \\
    & \le \triangle_i^{-\frac{1}{\varepsilon}} 2^\frac{2+3\varepsilon}{\varepsilon} v^{\frac{1}{\varepsilon}} c \log (TK) K T^{\frac{1}{M}},
\end{align*}
since $\tau_i(t_{M-1}^{\prime}) \le \tau_i^{\ast}$ , the total regret incurred by pulling arms in $\mathcal{I}_2$ is at most
\begin{align*}
    R_T(\pi^2) \mathcal{I}_2 & \le \sum_{i \in \mathcal{I}_2} \triangle_i^{-\frac{1}{\varepsilon}} 2^\frac{2+3\varepsilon}{\varepsilon} v^{\frac{1}{\varepsilon}} c \log (TK) K T^\frac{1}{M}.
\end{align*}

Add different regret parts together, we have
\begin{align*}
    \mathbb{E}\left[R_T\left(\pi^2\right)\right] \le
    C \sum_{i} \left( \frac{1}{\triangle_i}\right)^{\frac{1}{\varepsilon}} v^\frac{1}{\varepsilon} \log (TK) K T^\frac{1}{M},
\end{align*}
where $C > 0 $ is a constant only depend on $c$ .

\subsubsection*{Part III}

Finally, we examine the regret from $\min_{i \ne \star} \triangle_i$. With minor modifications to the former results in Part II, we obtain:
\begin{align*}
    \left \{  
    \begin{aligned}
        & R_T(\pi^2)_i \mathcal{I}_1
        \le 
        (\min \triangle_i)^{-\frac{1}{\varepsilon}} 2^\frac{2+3\varepsilon}{\varepsilon} v^{\frac{1}{\varepsilon}} c \log (TK) \cdot l^{\prime} \cdot \frac{K}{K-n_i} ,\\
        & R_T(\pi^2)_i \mathcal{I}_2
        \le 
        (\min \triangle_i)^{-\frac{1}{\varepsilon}} 2^\frac{2+3\varepsilon}{\varepsilon} v^{\frac{1}{\varepsilon}} c \log (TK) \cdot \frac{K}{t_{M-1}^{\prime}} \tau_i. 
    \end{aligned}
    \right.
\end{align*}
Then,
\begin{align*}
    \mathbb{E}\left[R_T\left(\pi^2\right)\right] & \le
    o(1) + R_T(\pi^1) \mathcal{I}_0 + R_T(\pi^1) \mathcal{I}_1 +R_T(\pi^1) \mathcal{I}_2 \\
    & \le o(1) + K^{\frac{\varepsilon}{1+\varepsilon}} T^{\frac{1}{M}} + (\min \triangle_i)^{-\frac{1}{\varepsilon}} 2^\frac{2+3\varepsilon}{\varepsilon} v^{\frac{1}{\varepsilon}} c \log (TK) T^{\frac{1}{M}} K \log K  \\
    & \quad + (\min \triangle_i)^{-\frac{1}{\varepsilon}} 2^\frac{2+3\varepsilon}{\varepsilon} v^{\frac{1}{\varepsilon}} c \log (TK) K T^{\frac{1}{M}} \\
    & \le
    C \left( \frac{1}{\min \triangle_i}\right)^{\frac{1}{\varepsilon}} v^\frac{1}{\varepsilon} \log(K) K \log (TK) T^\frac{1}{M},
\end{align*}
where $C > 0 $ is a constant only depend on $c$ .
\end{proof}

\subsection{Proof of Corollary \ref{cor:upper-M}}

\begin{proof}
    To achieve optimal regret upper bound, we compare batch number $M$ with batch number $T$ in Theorem \ref{thm:upper}. In short, we seek $M$  satisfying
    \begin{align*}
        \frac{\log K \left( K \log (TK) \right)^{\frac{\varepsilon}{1+\varepsilon}} T^{\frac{1}{1+\varepsilon-\varepsilon\left( \frac{\varepsilon}{1+\varepsilon} \right)^{M-1}}}}{\log K \left( K \log (TK) \right)^{\frac{\varepsilon}{1+\varepsilon}} T^{\frac{1}{1+\varepsilon-\varepsilon\left( \frac{\varepsilon}{1+\varepsilon} \right)^{T-1}}}} \le C,
    \end{align*}
    for come constant $C$.

    Rearrange the last inequality shows that
    \begin{align}
        T^{\frac{1}{1+\varepsilon-\varepsilon\left( \frac{\varepsilon}{1+\varepsilon} \right)^{M-1}}}
        \le
        C
        T^{\frac{1}{1+\varepsilon-\varepsilon\left( \frac{\varepsilon}{1+\varepsilon} \right)^{T-1}}}.
        \label{eq:proof-M}
    \end{align}
    Taking the logarithm on both sides gives that
    \begin{align*}
        \frac{1}{1+\varepsilon-\varepsilon\left( \frac{\varepsilon}{1+\varepsilon} \right)^{M-1}}
        \le 
        \frac{\log C}{\log T} 
        +
        \frac{1}{1+\varepsilon-\varepsilon\left( \frac{\varepsilon}{1+\varepsilon} \right)^{T-1}}.
    \end{align*}
    Equivalently, $M$ is constrained by 
    \begin{align*}
        \varepsilon \left( \frac{\varepsilon}{1+\varepsilon}\right)^{M-1} 
        \le
        \frac{\frac{\log C}{\log T} (1+\varepsilon) \left( 1+\varepsilon-\varepsilon\left( \frac{\varepsilon}{1+\varepsilon} \right)^{T-1} \right) + \varepsilon \left( \frac{\varepsilon}{1+\varepsilon} \right)^{T-1} }{\frac{\log C}{\log T} \left( 1+\varepsilon-\varepsilon\left( \frac{\varepsilon}{1+\varepsilon} \right)^{T-1} \right) + 1}.
    \end{align*}

    That is to say, for some $C>1$, lower bound of number of batches satisfies
    \begin{align}
        M
        \ge
        \log^{-1} \left( \frac{1+\varepsilon}{\varepsilon} \right)
        \log 
        \left(
        \frac{\varepsilon \left( 1+\varepsilon-\varepsilon\left( \frac{\varepsilon}{1+\varepsilon} \right)^{T-1} \right) \log C + \varepsilon \log T}{(1+\varepsilon) \left( 1+\varepsilon-\varepsilon\left( \frac{\varepsilon}{1+\varepsilon} \right)^{T-1} \right) \log C  + \varepsilon \left( \frac{\varepsilon}{1+\varepsilon} \right)^{T-1} \log T }
        \right) +1.
        \label{eq:upper-M-lower}
    \end{align}
    
    When $T$ is large enough, $\left( \frac{\varepsilon}{1+\varepsilon} \right)^{T-1} \log T = o (1)$.
    Number of batches $M$ satisfies Eq. (\ref{eq:upper-M-lower}) yields
    \begin{align*}
         M 
         =
         \Omega \left( 
         \log^{-1} \left( \frac{1+\varepsilon}{\varepsilon} \right)
         \log \log T
         \right).
     \end{align*}
\end{proof}

\newtheorem{lbdefinition}{Definition}[subsection]
\section{Lower Bounds}
\label{section:lower}
\subsection{Lower bound for the finite-arm setting}
Let $\Pi(M,T)$ be the set of policies with $M$ batches and horizon $T$, in this section, our objective is to lower bound the following \text{min-max} regrets and problem-dependent regrets under the batched setting:
\begin{align*}
    R_{\text{min-max}}=\inf_{\pi\in\Pi(M,T)}\sup_{\{\mu_i\}^K_{i=1}:\triangle_i\in[0,K^\frac{1}{\varepsilon}]}\mathbb{E}R_T(\pi),
\end{align*}
\begin{align*}
    R_{\text{pro-dep}}=\inf_{\pi\in\Pi(M,T)}\sup_{\triangle>0}\triangle^{\frac{1}{\varepsilon}}\cdot \sup_{\{\mu_i\}^K_{i=1}:\triangle_i\in\{0\}\cup[\triangle,K^\frac{1}{\varepsilon}]}\mathbb{E}R_T(\pi).
\end{align*}

In the formal discussion, the entire time horizon is splitted into $M$ batches represented by a grid $\mathcal{T}=\{0=t_0<t_1<\cdots<t_M=T\}$. Actually the grid belongs to following two categories: 

\begin{enumerate}
    \item Static grid (predetermined grid): the grid is determined before starting sampling any arms;
    \item Adaptive grid: for any $j\in[M]$, batch $t_j$ is determined after observations up to $t_{j-1}$. 
\end{enumerate}

We have the following lower bounds for these two settings: 
\begin{theorem}
    For an $M$-batched $K$-armed bandit problem with time horizon $T$ and any static grid of communication time points $\mathcal{T}$,
 the min-max regrets and problem-dependent regrets can be lower bounded as
    \begin{align*}
    R_{\text{min-max}} \ge \kappa \cdot K^{\frac{\varepsilon}{1+\varepsilon}}T^{\frac{1}{1+\varepsilon-\varepsilon(\frac{\varepsilon}{1+\varepsilon})^{M-1}}}
 , \quad \text{and}\quad 
    R_{\text{pro-dep}}\ge \kappa \cdot KT^{\frac{1}{M}},
\end{align*} 
for some constant $\kappa>0$ that does not depend on any parameter.
    \label{thm:lower-static}
\end{theorem} 

From the definition of problem-dependent regret, we can obtain lower bound of min-max regret with instance-dependent form under static grids:
$R_{\text{min-max}} \ge \kappa \left( \frac{1}{\min_{\triangle_i \neq 0} \triangle_i}  \right)^\frac{1}{\varepsilon} K T^\frac{1}{M}$. Next corollary focuses on the number of batches needed to achieve near-optimal
regret bound.

\begin{corollary}
    For an $M$-batched $K$-armed bandit problem with time horizon $T $ and any static grid of communication time points, the min-max regret has lower bound of order 
$\varOmega \left( K^{\frac{\varepsilon}{1+\varepsilon}}T^{\frac{1}{1+\varepsilon-\varepsilon(\frac{\varepsilon}{1+\varepsilon})^{M-1}}} \right)$ with only $\varOmega \left( \log^{-1}  \frac{1+\varepsilon}{\varepsilon} \log \log T \right)$ number of batches. While the min-max regret has lower bound of order 
$\varOmega \left( \left( \frac{1}{\min \triangle_i} \right)^\frac{1}{\varepsilon} K T^{\frac{1}{T}} \right)$ with $\varOmega \left( \log T \right)$ number of batches.
\label{cor:lower-M}
\end{corollary}

\begin{theorem}
   For an $M$-batched $K$-armed bandit problem with time horizon $T$ and any adaptive
   grid of communication time points, the \text{min-max} regrets and \text{pro-dep} regrets can be lower bounded as follows:
   \begin{align*}
        R_{\text{\text{min-max}}} \geq \kappa \cdot M^{-{\frac{3\varepsilon+1}{1+\varepsilon}}} \cdot K^{\frac{\varepsilon}{1+\varepsilon}} T^{\frac{1}{1+\varepsilon-\varepsilon\left( \frac{\varepsilon}{1+\varepsilon} \right)^{M-1}} }, \quad \text{and} \quad 
       R_{\text{pro-dep}}\geq \kappa \cdot M^{-3}
    \cdot K T^{\frac{1}{M}},
   \end{align*} 
    where $\kappa$ is a constant that that does not depend on any parameter. 
    \label{thm:lower-adaptive}
\end{theorem}

\subsection{Lower bound for the Lipschitz setting}


The lower bounds that represent the hardness of such problems is stated below in Theorem \ref{thm:lower-lips} and Corollary \ref{cor:lower-lips-M}. 

\begin{theorem}
    Consider a $T$-step Lipschitz bandit problem where the reward distributions have $(1 + \varepsilon)$-th moment bounded by $v< \infty$. 
    Suppose that the algorithm are allowed to collect rewards at 
    a predetermined grid of $M$ communication time points. 
    Then the min-max regret is lower bounded as follows:
    \begin{align*}
        R_{\text{min-max}} \ge  \kappa \varepsilon^\frac{1}{2} T^\frac{1-\frac{1}{d_z+1+\frac{1}{\varepsilon}}}{1-(\frac{1}{d_z+1+\frac{1}{\varepsilon}})^M}. 
    \end{align*}
    where $\kappa > 0$ is a numerical constant independent of $T$. 
    If instead the algorithm is allowed to collect rewards at 
    an adaptive grid of $M$ time points. 
    Then the min-max regret is lower bounded as follows:
    \begin{align*}
        R_{\text{min-max}} \ge 
        \kappa \frac{1}{M^2} T^{\frac{1-\frac{1}{d_z+1+\frac{1}{\varepsilon}}}{1-\left( \frac{1}{d_z+1+\frac{1}{\varepsilon}}\right)^M}},
    \end{align*}
    where $\kappa > 0$ is a numerical constant independent of $T$.
    \label{thm:lower-lips}
\end{theorem}

\begin{corollary}
    Consider a $T$-step Lipschitz bandit problem where the reward distributions have $(1 + \varepsilon)$-th moment bounded by $v$. 
    Suppose that the algorithm are allowed to collect rewards at 
    a predetermined grid of $M$ time points. 
    Then the min-max regret has lower bound of order $\varOmega \Bigg( T^{\frac{1-\frac{1}{d_z+1+\frac{1}{\varepsilon}}}{1-\left( \frac{1}{d_z+1+\frac{1}{\varepsilon}}\right)^T}} \Bigg) = \varOmega \left( T^{ \frac{d_z + \frac{1}{\varepsilon}}{d_z + 1 + \frac{1}{\varepsilon} } } \right) $ with only ${\varOmega} \left( \log^{-1} \left( d_z+1+\frac{1}{\varepsilon} \right) \log \log T \right)$ number of batches. 
    \label{cor:lower-lips-M} 
\end{corollary}

As before, the analysis of the finite-arm setting is developed in the following section, where we present our instance constructions for both the static and adaptive grid cases, while the corresponding proofs for the Lipschitz setting are provided in the Appendix.

\section{Lower bound analysis for the finite-arm setting}
\label{app:lower}





In the formal sections, our analysis is based on static grid. But noting the fact that the adaptive grid is much more powerful and practical than static grid to a certain extent. So in this section, we will  both focus on the lower bound analysis under static grid and adaptive grid. Our main results are shown as \textbf{Theorem \ref{thm:lower-static}} and \textbf{Theorem \ref{thm:lower-adaptive}}.

\subsection{Analysis for the static grid case}

For the minimax lower bound, we construct a family of problem instances where no algorithm can perform well uniformly across all cases. We then describe the specific instances used to establish the lower bound. 
Consider $\triangle_0 \in (0,\frac{1}{2})$ and let $\gamma = (2\triangle_0)^{\frac{1}{\varepsilon}}$, and define the following distributions parametrized by $\triangle \in (0,\frac{1}{2})$: 
\begin{align*} 
    \nu(\triangle)=(1-\gamma^{1+\varepsilon}+\triangle_{0} \gamma+\triangle\gamma)\delta_{0}+(\gamma^{1+\varepsilon}-\triangle_{0}\gamma+\triangle\gamma)\delta_{\frac{1}{\gamma}} . 
\end{align*}
where $ \delta_z $ is the Dirac delta at $z$. It is straightforward to verify that $ \gamma^{1+\varepsilon}-\triangle_{0}\gamma+\triangle\gamma \in (0,1) $, which implies that $\nu(\triangle)$ is a valid distribution. In addition, it holds that 
\begin{align*} 
    \E [ \nu (\triangle) ] = \triangle_0 + \triangle 
    \quad \text{and} \quad 
    \E [ | \nu (\triangle) - \E [ \nu (\triangle) ] |^{1 + \varepsilon} ] < \infty.  
\end{align*} 

Based on distributions $ \nu (\triangle) $, we construct problem instances: $ \forall j \in [M - 1], \forall k \in [K - 1], $ 
\begin{align*}
    P_{j,k} = \nu(0) \otimes \cdots \otimes \nu(0) \otimes \nu(\triangle_j + \triangle_M) \otimes \nu(0) \otimes \cdots \otimes \nu(0) \otimes \nu(\triangle_M),
\end{align*}
and 
    $P_M = \nu(0) \otimes \cdots \otimes \nu(0) \otimes \nu(\triangle_M), $
where 
\begin{align*} 
    \triangle_{j} = \frac{K^{\frac{\varepsilon}{1+\varepsilon}} T^{-\frac{\varepsilon}{1+\varepsilon}\left(\frac{1+\varepsilon-\varepsilon\left( \frac{\varepsilon}{1+\varepsilon} \right)^{j-2}}{1+\varepsilon-\varepsilon\left( \frac{\varepsilon}{1+\varepsilon} \right)^{M-1}}\right)}}{(12\sqrt{ 6\cdot2^\frac{1}{\varepsilon} }M)^{\frac{2\varepsilon}{1+\varepsilon}}2^{\frac{3\varepsilon+1}{1+\varepsilon}}} 
    \quad \text{for } \forall j\in[M] . 
\end{align*} 

For any policy, there exists some instance (among such instances) that will trick the policy into bad performance. The detailed proofs are deferred to the Appendix.

The proof of Theorem \ref{thm:lower-static} relies on the following lemma. We defer the formal proof, to Appendix \ref{app:lem}.
\begin{lemma}
    For any grid $0=t_0 < t_1 < t_2 < t_3 < \cdots < t_M=T$ and the smallest gap $\triangle\in (0,K^{\varepsilon/(1+\varepsilon)}]$, the following  minimax lower bound holds for any policy $\pi$,
    \begin{align*}
        \sup_{\{\mu_i\}^K_{i=1}:\triangle_i\in\{0\}\cup[\triangle,K^{\varepsilon/(1+\varepsilon)}]} \mathbb{E}R_T(\pi)\geq \triangle \cdot \sum^M_{j=1} \frac{t_j-t_{j-1}}{4} \exp{\left(- \frac{C 3^{\frac{1}{\varepsilon}} \triangle^{\frac{1+\varepsilon}{\varepsilon}}t_{j-1}}{K-1}\right)},
    \end{align*}
    where $C$ is a numerical constant.
    \label{lem:lower-static}
\end{lemma}

By the definition of \text{min-max} regret and \text{pro-dep} regret , we choose $\triangle=\triangle_j=\left(\frac{K-1}{ 3^{\frac{1}{\varepsilon}} (t_{j-1}+1)}\right)^{\frac{\varepsilon}{1+\varepsilon}}$ in \textbf{Lemma} \ref{lem:lower-static} yields that
\begin{align*}
    R_{\text{min-max}}\geq \kappa \cdot K^{\frac{\varepsilon}{1+\varepsilon}}\max_{j\in [M]}\frac{t_j}{(t_{j-1}+1)^{\frac{\varepsilon}{1+\varepsilon}}}
\end{align*}
\begin{align*}
    R_{\text{pro-dep}}\geq \kappa \cdot K\max_{j\in [M]}\frac{t_j}{t_{j-1}+1}
\end{align*} 
which implies the lower bound in Theorem \ref{thm:lower-static}.


\begin{proof}[Proof of Corollary \ref{cor:lower-M}]
We focus on the two forms of mini-max regret under static grid, including instance-independent form and instance-dependent form.

In regard to instance-independent form:
\begin{align*}
    R_{\text{min-max}} \ge \kappa 
    K^{\frac{\varepsilon}{1+\varepsilon}} T^{\frac{1}{1+\varepsilon-\varepsilon\left( \frac{\varepsilon}{1+\varepsilon} \right)^{M-1}} },
\end{align*}
we hope to find $M$ such that
\begin{align*}
    K^{\frac{\varepsilon}{1+\varepsilon}} T^{\frac{1}{1+\varepsilon-\varepsilon\left( \frac{\varepsilon}{1+\varepsilon} \right)^{M-1}} }
    \le 
    C K^{\frac{\varepsilon}{1+\varepsilon}} T^{\frac{1}{1+\varepsilon-\varepsilon\left( \frac{\varepsilon}{1+\varepsilon} \right)^{T-1}} },
\end{align*}
for some constant $C$. At this point, it was discovered that this form was exactly the same as Eq. (\ref{eq:proof-M}). From its calculation, we get
\begin{align*}
    M 
    =
    \varOmega \left( 
    \log^{-1} \left( \frac{1+\varepsilon}{\varepsilon} \right)
    \log \log T
    \right).
\end{align*}

With respect to instance-dependent, our results is 
\begin{align*}
    R_{\text{min-max}} \ge \kappa
    \left( \frac{1}{\min \triangle_i} \right)^\frac{1}{\varepsilon} K T^\frac{1}{M}.
\end{align*}
For $M$ satisfying $T^\frac{1}{M} \le C T^\frac{1}{T}$ for some constant $C$ implies
\begin{align*}
    M = \varOmega (\log T).
\end{align*}
\end{proof}

\subsection{Analysis for the adaptive grid case}

We first focus on proving the lower bound for the \text{min-max} regret, as this case most clearly illustrates the fundamental property of regret. The proof of the \text{pro-dep} regret is totally analogous.
Let's considering the following time $T_{1},T_{2},...,T_{M}\in[1,T]$
and gaps $\triangle_{1},\triangle_{2},...,\triangle_{M}\in(0,K^{\frac{\varepsilon}{1+\varepsilon}}]$ with
\begin{align*}
    T_{j}=\lfloor T^{\frac{1+\varepsilon-\varepsilon\left( \frac{\varepsilon}{1+\varepsilon} \right)^{j-1}}{1+\varepsilon-\varepsilon\left( \frac{\varepsilon}{1+\varepsilon} \right)^{M-1}}}\rfloor, \quad
    \text{and} \quad
    \triangle_{j}=\frac{K^{\frac{\varepsilon}{1+\varepsilon}} T^{-\frac{\varepsilon}{1+\varepsilon}\left(\frac{1+\varepsilon-\varepsilon\left( \frac{\varepsilon}{1+\varepsilon} \right)^{j-2}}{1+\varepsilon-\varepsilon\left( \frac{\varepsilon}{1+\varepsilon} \right)^{M-1}}\right)}}{(12\sqrt{\alpha}M)^{\frac{2\varepsilon}{1+\varepsilon}}2^{\frac{3\varepsilon+1}{1+\varepsilon}}}
    \quad \text{for } \forall j\in[M],
\end{align*}
where $\alpha=6\cdot2^\frac{1}{\varepsilon}$.

For the minimax lower bound proof, we construct a set of problem instances such that no algorithm can uniformly perform well on these instances. Next we describes the problem instances used for proving the lower bound. Consider $\Delta_0 \in (0,\frac{1}{2})$ and let $\gamma = (2\Delta_0)^{\frac{1}{\varepsilon}}$, and defined the following distributions parametrized by $\Delta \in (0,\frac{1}{2})$: 
\begin{align*} 
    \nu(\triangle)=(1-\gamma^{1+\varepsilon}+\triangle_{0} \gamma+\triangle\gamma)\delta_{0}+(\gamma^{1+\varepsilon}-\triangle_{0}\gamma+\triangle\gamma)\delta_{\frac{1}{\gamma}} . 
\end{align*}
where $ \delta_z $ is the Dirac delta at $z$. It is straightforward to verify that $ \gamma^{1+\varepsilon}-\triangle_{0}\gamma+\triangle\gamma \in (0,1) $, which implies that $\nu(\triangle)$ is a valid distribution. In addition, it holds that 
\begin{align*} 
    \E [ \nu (\Delta) ] = \Delta_0 + \Delta 
    \quad \text{and} \quad 
    \E [ | \nu (\Delta) - \E [ \nu (\Delta) ] |^{1 + \varepsilon} ] < \infty.  
\end{align*}


Let $\mathcal{T} = \{t_1, \cdots, t_M\}$ be any adaptive grid, and $\pi$ be any policy under the grid $\mathcal{T}$. For each $j \in [M]$, we define the event $A_j = \{t_{j-1} < T_{j-1}, t_j \geq T_j\}$ under policy $\pi$ with the convention that $t_0 = 0, t_M = T$. The union of events $A_1, \cdots, A_M$ forms a cover of the entire sample space. Under any admissible probability distribution $\mathbb{P}$, $\sum_{i=1}^M \mathbb{P}(A_i) \ge \mathbb{P}(\cup_{i=1}^M A_i)\ge 1$ holds.
We are interested in the following family of reward distributions:
\begin{align*}
    P_{j,k} = \nu(0) \otimes \cdots \otimes \nu(0) \otimes \nu(\triangle_j + \triangle_M) \otimes \nu(0) \otimes \cdots \otimes \nu(0) \otimes \nu(\triangle_M),& \\
    \forall j \in [M - 1], \forall k \in [K - 1],&
\end{align*}
where the $k$-th component of $P_{j,k}$ has a non-zero mean. For $j = M$, we define
\begin{align*}
    P_M = \nu(0) \otimes \cdots \otimes \nu(0) \otimes \nu(\triangle_M).
\end{align*}

Note that in this family of reward distributions, $P_{j,k}$ and $P_M$ only differs in the $k$-th component, which is similar to the distributions in our proof for static grid.

By our observations,if $A_j$ occurs with a fairly large probability, the result in total regret will be large as well. So,we will be interested in the following probabilities:
\begin{align*}
    p_j = \frac{1}{K - 1} \sum_{k=1}^{K-1} P_{j,k}(A_j), \quad j \in [M - 1], \quad p_M = P_M(A_M),
\end{align*}
where $P_{j,k}(A)$ denotes the probability of the event $A$ given the reward distribution $P_{j,k}$ under the policy $\pi$. Then, we have the following lemmas.

\begin{lemma}
    \label{lem:lower-pro}
    The following inequality holds: $\sum_{j=1}^{M}p_{j}\geq{}\frac{1}{2}.$
\end{lemma}

\begin{lemma}
    \label{lem:lower-mini}
    If $p_{j}\geq{\frac{1}{2M}}$ for some $j\in[M]$, then we have
    \begin{align*}\sup_{\{\mu_i \}_{i=1}^{K} : \triangle_{i} \leq {}K^{\frac{\varepsilon}{1+\varepsilon}}} \mathbb{E}[R_{T}(\pi)] \ge \kappa \cdot M^{-{\frac{3\varepsilon+1}{1+\varepsilon}}}
    \cdot K^{\frac{\varepsilon}{1+\varepsilon}} T^{\frac{1}{1+\varepsilon-\varepsilon\left( \frac{\varepsilon}{1+\varepsilon} \right)^{M-1}}},\end{align*}
    where $ \kappa >0$ is a  constant independent of any parameter.
\end{lemma} 

For complete technical details of the lemmas, see Appendix \ref{app:lem}. Theorem \ref{thm:lower-adaptive} follows directly from these lemmas through a straightforward derivation.

\section{Experiments}
\label{experiments}

\begin{figure}[!h]
    \centering
    \includegraphics[scale = 0.45]{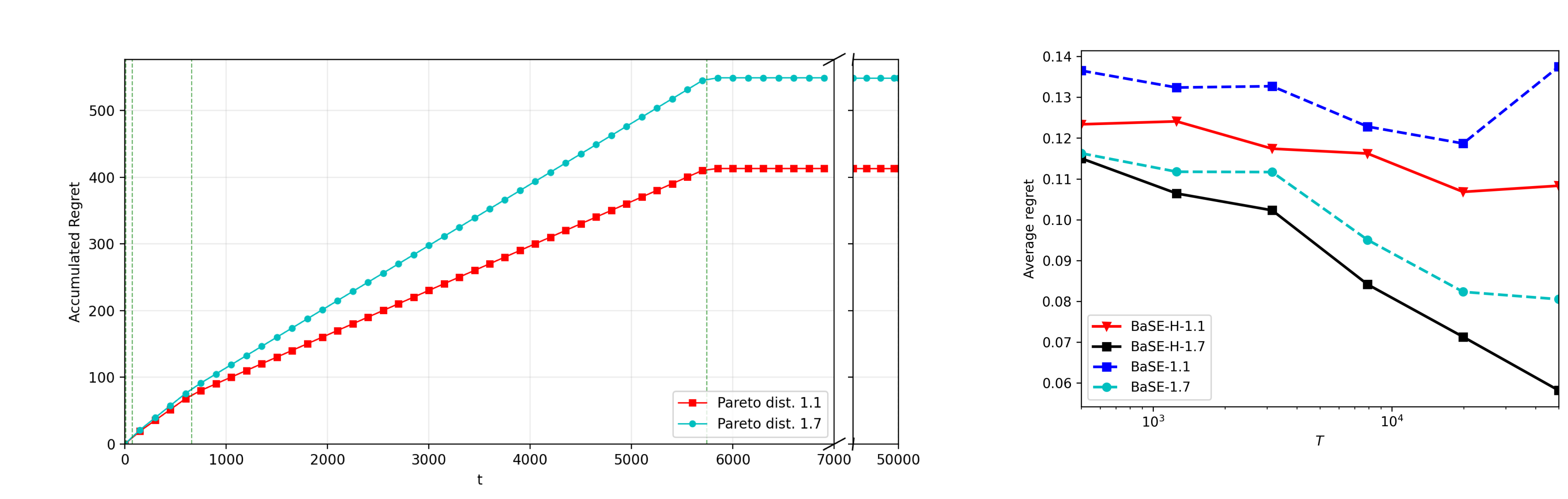}
    \caption{Empirical regret of the BaSE-H algorithm. The left panel plots the accumulated regret of BaSE-H algorithm over time points $t = 1$ to $T$ under two different Pareto noise distributions with scale parameters $1.1$ and $1.7$. The green vertical dashed lines indicate the communication time points $\mathcal{T}$. The right panel plots the average regret against the time horizon $T$ for two algorithms under the Pareto($1.1$) and Pareto($1.7$) distributions. These two algorithms are BaSE-H algorithm in our work and BaSE algorithm in \citep{gao2019batched}. Here the accumulated regret refers to the increased regret incurred by our algorithm over time, and the average regret at time horizen $T$ is $R_{T}(\pi)/T$.}
    \label{fig:finite}
\end{figure}

In this section, we present experimental results illustrating the performance of our algorithms under different settings, along with a comparison to existing batched bandit algorithms \citep{gao2019batched,feng2022lipschitz}. 
In the finite-arm setting, we set $T = 5 \times 10^4$, with the grid defined under the instance-dependent case as in \eqref{eq:grid-dep}. The expected reward of the optimal arm is $\mu_\star = 1$, while for all suboptimal arms $\mu_i = 0.8$ ($i \neq \star$). We consider two heavy-tailed Pareto distributions with shape parameters $1.1$ and $1.7$, which have finite $(1+\varepsilon)$-th moments for $\varepsilon < 0.1$ and $\varepsilon < 0.7$, respectively.
Fig. \ref{fig:finite} 
presents the empirical regret under different noise distributions, together with a comparison to the BaSE algorithm \citep{gao2019batched}, which does not account for heavy-tailed noise.
The parameters for the left panel (accumulated regret) are $K = 6$ and $M = 5$, while those for the right panel (average regret) are $K = 3$ and $M = 3$.
Our BaSE-H Algorithm 1 outperforms BaSE under heavier-tailed rewards, and heavier-tailed distributions have larger regret as expected.

\begin{figure}[!h]
    \centering
    \includegraphics[scale = 0.46]{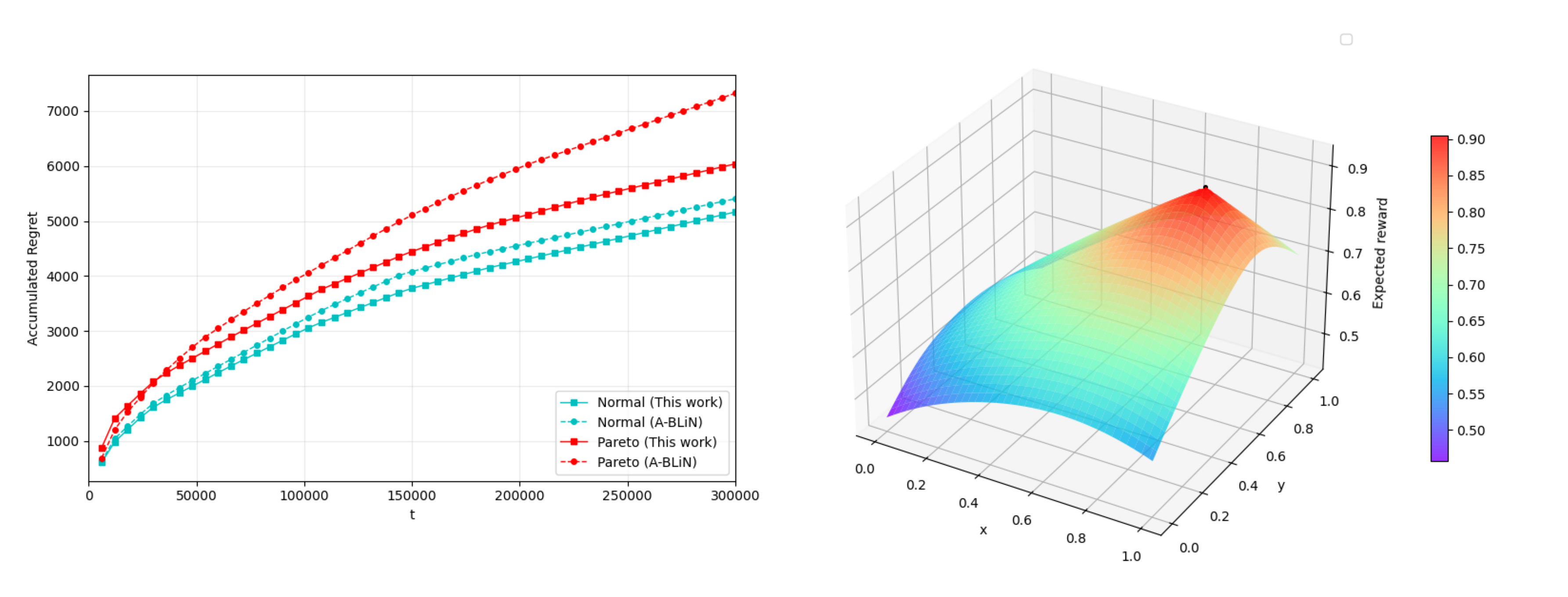}
    \caption{Accumulated regret of the algorithm \ref{alg:algorithm-lips}. The left panel plots the accumulated regret against time points $t$ from 1 to $T$ under the Pareto($1.7$) and Normal $\mathcal{N}(0,1)$ distributions. The right panel shows the diagram of expected reward under reward function: $1- 0.5 \times \|(x - 0.8, y - 0.7) \|_2 - 0.1 \times \| (x - 0.1, y - 0.2) \|_2$.}
    \label{fig:Lips}
\end{figure}

In the Lipschitz bandit setting, we set arm space $[ 0, 1 ]^2$, and time horizen $T = 3 \times 10^5$. The expected reward function is $\mu(x, y) = 1- 0.5 \times \|(x - 0.8, y - 0.7) \|_2 - 0.1 \times \| (x - 0.1, y - 0.2) \|_2$, with its optimal arm at $(0.8, 0.7)$. 
Fig. \ref{fig:Lips} shows the accumulated regret under two noise settings: Gaussian noise and heavy-tailed noise. The heavy-tailed noise is sampled from a normalized Pareto distribution with shape parameter $1.7$, whose $1.5$-th centered moment is finite and less than $1$.
We evaluate our algorithm against A-BLiN \cite{feng2022lipschitz} in both scenarios. Under Gaussian noise, both algorithms achieve comparable performance. However, in the presence of heavy-tailed noise, the existing method performs poorly due to an erroneous elimination step. As indicated by the red cube-marked line in Fig. \ref{fig:Lips}, our algorithm benefits from thorough sampling in the early stage, enabling it to perform well as $t$ increases.
The code can be found at \url{https://github.com/YunluShu/Batched-Bandits-with-Heavy-Tailed-Rewards}.

\section{Conclusion} 

This paper addresses the critical problem in batched multi-armed bandit (MAB) problems by introducing robust algorithms tailored for heavy-tailed reward distributions, a common yet understudied scenario in real-world applications like clinical trials. 
The proposed methods are further extended to batched bandit algorithms for heavy-tailed rewards in Lipschitz settings. For both the multi-armed and Lipschitz settings, this paper provides analyses of regret upper and lower bounds, as well as the corresponding required number of batches.
The study reveals intriguing phenomena regarding the interplay between tail heaviness and communication patterns in batched settings.

\bibliographystyle{plain} 
\bibliography{references} 

\clearpage

\appendix

\section{Analysis with Arm-Specific Tail Parameters under Finite-arm Setting}
\label{app:arm-specific}

In this section, we extend our analysis to the setting where each arm $i$ possesses a distinct heavy-tailed parameter $\varepsilon_i$. To accommodate the arm-specific parameters in the static grid construction, we first define the global minimum parameter $\varepsilon'$ and the corresponding number of batches $l$.

Let $\varepsilon' = \min_{i \in \mathcal{X}} \{\varepsilon_i\}$ and $\varepsilon'_i = \min\{\varepsilon_i, \varepsilon_{\star}\}$, where $\varepsilon_{\star}$ is the heavy-tailed parameter of the optimal arm $\star$. We define the number of batches $l$ by substituting $\varepsilon$ with $\varepsilon'$ in the original grid formulation:
\begin{equation}
    l = T^{\frac{1}{1+\varepsilon' - \varepsilon'\left( \frac{\varepsilon'}{1+\varepsilon'} \right)^{M-1}}}.
\end{equation}

And we can modify the elimination condition in Line 7 of Algorithms \ref{alg:base} (BaSE-H) as follows:

Replace the original condition
$$\hat{\mu}_{\max}(\tau(t_m)) - \hat{\mu}_i(\tau(t_m)) \geq 2^{\frac{1+2\varepsilon}{1+\varepsilon}} v^{\frac{1}{1+\varepsilon}} \left(\frac{c \log(TK)}{\tau(t_m)}\right)^{\frac{\varepsilon}{1+\varepsilon}}$$
with the arm-specific version:
$$\hat{\mu}_{\max}(\tau(t_m)) - \hat{\mu}_i(\tau(t_m)) \geq  2^{\frac{1+2\varepsilon'_i}{1+\varepsilon'_i}} v^{\frac{1}{1+\varepsilon'_i}} \left(\frac{c \log(TK)}{\tau(t_m)}\right)^{\frac{\varepsilon'_i}{1+\varepsilon'_i}}.$$

By utilizing $\varepsilon' \text{and } \varepsilon'_i$, we ensure that the concentration inequality required for Lemma \ref{lem:upper} holds uniformly. Consequently, the subsequent derivation of the upper bound follows the structure of the uniform case but with arm-dependent constants.

  And the critical threshold $\tau_i^*$ is changed to:
\begin{equation}
    \tau_i^* \triangleq 2^{\frac{2+3\varepsilon'_i}{\varepsilon'_i}} v^{\frac{1}{\varepsilon'_i}} c \log(TK) \left(\frac{1}{\Delta_i}\right)^{\frac{1+\varepsilon'_i}{\varepsilon'_i}}.
\end{equation}

Let $\pi^1$ be the policy equipped with the modified communication time points $\mathcal{T}^{1'}$ (defined based on $l$ above) and the modified elimination condition. The regret upper bound is stated as follows:

\begin{theorem}[Regret Bound with Arm-Specific Parameters]
\label{thm:arm_specific}
Suppose each arm $i$ satisfies the $(1+\varepsilon_i)$-th moment condition with bound $v$. Under the policy $\pi^1$ described above, the expected cumulative regret is bounded by:
\begin{equation}
    \mathbb{E}[R_T(\pi^1)] \le \sum_{i \in \mathcal{X}} 2^{\frac{2+3\varepsilon'_i}{1+\varepsilon'_i}} v^{\frac{1}{1+\varepsilon'_i}} (c \log(TK) K)^{\frac{\varepsilon'_i}{1+\varepsilon'_i}} T^{\frac{1}{1+\varepsilon' - \varepsilon'\left( \frac{\varepsilon'}{1+\varepsilon'} \right)^{M-1}}}.
\end{equation}
\end{theorem}

\begingroup

\begin{proof}
We now present a brief sketch of the proof for the grid setting with arm-specific parameters; the proofs for other settings are analogous.

For arms in $\mathcal{I}_1$, given the condition $\tau_i^* > \tau_i \ge \frac{t_{m_i-1}}{K} - 1$, we derive the following inequality governing the batch size:
\begin{equation}
    2^{\frac{2+3\varepsilon'_i}{\varepsilon'_i}} v^{\frac{1}{\varepsilon'_i}} c \log(TK) \left(\frac{1}{\Delta_i}\right)^{\frac{1+\varepsilon'_i}{\varepsilon'_i}} \ge \frac{t_{m_i-1}}{K}.
\end{equation}
Hence, the total regret incurred by pulling an arm $i \in \mathcal{I}_1$ is at most:
\begin{align}
    R_T(\pi^1)_i &\le \Delta_i \cdot \tau_i \le \Delta_i \frac{t_{m_i}}{K - n_i} \nonumber \\
    &\le 2^{\frac{2+3\varepsilon'_i}{1+\varepsilon'_i}} v^{\frac{1}{1+\varepsilon'_i}} (c \log(TK) \cdot K)^{\frac{\varepsilon'_i}{1+\varepsilon'_i}} \cdot \frac{1}{(t_{m_i-1})^{\frac{\varepsilon'_i}{1+\varepsilon'_i}}} \cdot \frac{t_{m_i}}{K - n_i} \nonumber \\
    &\le 2^{\frac{2+3\varepsilon'_i}{1+\varepsilon'_i}} v^{\frac{1}{1+\varepsilon'_i}} (c \log(TK) K)^{\frac{\varepsilon'_i}{1+\varepsilon'_i}} l \frac{1}{K - n_i}.
\end{align}
Summing over all arms in $\mathcal{I}_1$, we obtain the cumulative regret for Part I:
\begin{equation}
    R_T(\pi^1)_{\mathcal{I}_1} \le \sum_{i \in \mathcal{I}_1} R_T(\pi^1)_i \le \sum_{i \in \mathcal{I}_1} 2^{\frac{2+3\varepsilon'_i}{1+\varepsilon'_i}} v^{\frac{1}{1+\varepsilon'_i}} (c \log(TK) K)^{\frac{\varepsilon'_i}{1+\varepsilon'_i}} l.
\end{equation}

For the arms in $\mathcal{I}_2$, the regret analysis yields the following bound for a single arm $i$:
\begin{align}
    R_T(\pi^1)_i &\le \Delta_i \cdot \tau_i \nonumber \\
    &\le \tau_i \cdot 2^{\frac{2+3\varepsilon'_i}{1+\varepsilon'_i}} v^{\frac{1}{1+\varepsilon'_i}} (c \log(TK) \cdot K)^{\frac{\varepsilon'_i}{1+\varepsilon'_i}} \cdot \frac{l}{T}.
\end{align}

Since the total pulls are bounded by the horizon, i.e., $\sum_{\mathcal{I}_2} \tau_i \le T$, the total regret incurred by pulling arms in $\mathcal{I}_2$ is at most:
\begin{equation}
    R_T(\pi^1)_{\mathcal{I}_2} \le \sum_{i \in \mathcal{I}_2} 2^{\frac{2+3\varepsilon'_i}{1+\varepsilon'_i}} v^{\frac{1}{1+\varepsilon'_i}} (c \log(TK) K)^{\frac{\varepsilon'_i}{1+\varepsilon'_i}} l.
\end{equation}

Finally, we sum up those regrets to obtain our upper bound for arm specific analysis:
\begin{equation}
    R_T(\pi^1) \le \sum_{i } 2^{\frac{2+3\varepsilon'_i}{1+\varepsilon'_i}} v^{\frac{1}{1+\varepsilon'_i}} (c \log(TK) K)^{\frac{\varepsilon'_i}{1+\varepsilon'_i}} l.
\end{equation}

\end{proof}

\endgroup 

\section{Upper bound analysis for the Lipschitz setting}
\label{app:lips} 

\subsection{Notations}

For self-containedness of the Appendix, we first recap some notations used in this setting. For the Lipschitz bandit problem, the arm set is a compact doubling metric space $( \mathcal{X},d_{\mathcal{X}})$. Here we take $\mathcal{X} = [0,1]^d$ and $d_{\mathcal{X}} = \| \cdot \|_\infty$. 
Pulling an arm in cube means choosing its center.

The expected reward $\mu$ on $A$ is a 1-Lipschitz function with respect to the metric $d_{\mathcal{X}}$, that is to say, $|\mu(x) - \mu(y)| \le \| x - y\|_\infty$ for any $x, y \in \mathcal{X}$. The corresponding definition for the optimal arm is $\mu_\star = max_{x \in \mathcal{X}} \mu(x)$. For any arm $x$, gap is $\triangle_x = \mu_\star - \mu(x)$. 

The zooming dimension is then defined as $d_z$ and the zooming constant is $C_z$.

At time $t$, no matter which $x$ we sample, reward $\nu(x)$ has mean $\mu(x)$ and bounded $1+\varepsilon$ moments with bound $v$. Due to robustness, if we choose domain $C \subset [0,1]^d$ and sample from $C$ several times, mean estimators of these rewards can estimate expected reward of this domain.

\subsection{Proof of Theorem \ref{thm:upper-lips}}

For Lipschitz bandit, we propose an algorithm based on cube partitioning. The arm space is initially divided into uniform hypercubes, which are iteratively refined through batches. Each batch uniformly samples active cubes, and eliminates cubes with low reward. Then we partition remaining cubes into finer subcubes and advance to the next batch.

Recall the edge-length within each batch is specified by Definition \ref{def:lips-rm}. 


At the beginning of each batch $m$, a collection of subsets of the arm space $A_m$ is constructed. This collection $A_m$ consists of standard cubes with the same edge-length $r_m$. Then each cube in $A_m$ is played $n_m$ times. We choose sample number $\{ n_m \}$ according to $\{r_m\}$. Specifically,
\begin{align*}
    n_m = 3c v^{\frac{1}{\varepsilon}} \log T \cdot r_m^{-\frac{1+\varepsilon}{\varepsilon}} \quad \text{for } m=1,\cdots.
\end{align*}

To ensure feasible subdivisions at each batch, the more precise operation involves rounding $\frac{r_m}{r_{m+1}}$. Since this rounding operation preserves the original regret order, we proceed with the $\frac{r_m}{r_{m+1}}$ in our analysis.
The detailed process is revealed in \textbf{Algorithm} \ref{alg:algorithm-lips}.


Similar to the proof of \textbf{Theorem} \ref{thm:upper}, we first estimate the probability that the best arm $\star$ is not removed and our mean estimators are reliable. Let $B$ be the event that the best arm $\star$ is not eliminated throughout time horizon $T$. Define
\begin{align*}
    E:=\left\{\left|\mu(x)-\hat{\mu}_C(n_m)\right| \leq 2 r_m, \forall C \in \mathcal{A}_m, \forall x \in C\right\}.
\end{align*}

\begin{lemma}
    Define event $\mathcal{E}$ as $B \cap E$. Then, it holds that $\mathbb{P}(\mathcal{E}) \geq 1-2 T^{-6}$.
    \label{lem:upper-lips}
\end{lemma}

We proof this lemma in Appendix \ref{app:lem}. With the lemma, finally we carry out the proof of \textbf{Theorem} \ref{thm:upper-lips} below.

\begin{proof}[Proof of Theorem \ref{thm:upper-lips}]
    Since the expected reward under $\mathcal{E}^c$ is $o(1)$, here we only need to calculate under event $\mathcal{E}$.
    
    We first show that the cubes that survive after each batch own high rewards. After $m-1$-th batch, we consider $x$ within $A_{m-1}^+$, or equally $x$ within $A_m$. We use $C$ to donate the cube containing $x$ in $A_{m-1}$ and $C_\star$ to donate the cube containing $\star$. Our interest focuses on $\triangle_x$,
    \begin{align*}
        \triangle_x = \mu_\star - \mu(x)
        & \le \hat{\mu}_{C_\star}(n_{m-1}) +2 r_{m-1} - ( \hat{\mu}_{C}(n_{m-1}) -2 r_{m-1} )\\
        & \le \hat{\mu}_{C_\star}(n_{m-1})  - \hat{\mu}_{C}(n_{m-1}) +4 r_{m-1}.
    \end{align*}
    By \textbf{Algorithm} \ref{alg:algorithm-lips}, $C$ and $C_\star$ are remaining subcubes after batch $m-1$, hence $| \hat{\mu}_{C_\star}(n_{m-1})  - \hat{\mu}_{C}(n_{m-1}) | \le 4 r_{m-1}$. We can conclude that for any $x$ within $A_{m-1}^+$ or $A_m$,
    \begin{align}
        \triangle_x \le 8 r_{m-1} = 8 r_m 2^{c_m}.
        \label{eq:upper-lips-tri-x}
    \end{align}

    Now we turn to upper bound regret $R_T$. Let $R_m$ donate the expected regret from the $m$-th batch. During batch $m$, algorithm samples each subcubes in $A_m$ $n_m$ times, and each sample has loss upper bounded by $8 r_{m-1}$ from Eq.(\ref{eq:upper-lips-tri-x}). The total regret of the $m$-th batch is
    \begin{align*}
        R_m \le |A_m| \cdot n_m \cdot 8 r_{m-1} = 24cv^\frac{1}{\varepsilon} |A_m| r_m^{-\frac{1+\varepsilon}{\varepsilon}} r_{m-1} \log T.
    \end{align*}
    Due to division principle, $|A_m| = |A_{m-1}^+| \left( \frac{r_{m-1}}{r_m} \right)^d$, and here $A_{m-1}^+$ is a subset of $S(8 r_{m-1})$.
    In retrospect, zooming dimension is $d_z$. Count of $|A_{m-1}^+|$ is no more than $C_z r_{m-1}^{-d_z}$. Following from these statement,
    \begin{align*}
        R_m & \le 24cv^\frac{1}{\varepsilon} C_z r_{m-1}^{-d_z} \left( \frac{r_{m-1}}{r_m} \right)^d r_m^{-\frac{1+\varepsilon}{\varepsilon}} r_{m-1} \log T \\
        & = 24cv^\frac{1}{\varepsilon} C_z \log T r_{m-1}^{-d_z+d+1} r_m^{-d-\frac{1+\varepsilon}{\varepsilon}} .
    \end{align*}
    Adopt \textbf{Definition} \ref{def:lips-rm}, we have
    \begin{align*}
        R_m  \le 24cv^\frac{1}{\varepsilon} C_z \log T \cdot 2^{(\sum_{i=1}^{m-1} c_i) \left( d_z +\frac{1}{\varepsilon}  \right) +c_m \left( d+\frac{1+\varepsilon}{\varepsilon}  \right)}.
    \end{align*}
    We define $C_m = (\sum_{i=1}^{m-1} c_i) \left( d_z +\frac{1}{\varepsilon}  \right) +c_m \left( d+\frac{1+\varepsilon}{\varepsilon}  \right)$. Using $\eta = \frac{d+1-d_z}{d+1+\frac{1}{\varepsilon}}$, the recursion formula is $C_m = C_{m-1} + c_{m-1}(d_z -1 -d) + c_m\left( d+\frac{1+\varepsilon}{\varepsilon}  \right) = C_{m-1} = C_1 = c_1 \left( d+\frac{1+\varepsilon}{\varepsilon}  \right)$. So,
    \begin{align*}
        R_m  & \le 24cv^\frac{1}{\varepsilon} C_z \log T \cdot 2^{c_1 \left( d+\frac{1+\varepsilon}{\varepsilon}  \right)} \\
        & = 24cv^\frac{1}{\varepsilon} C_z \log T \cdot \left( \frac{T}{\log T} \right)^\frac{d_z +\frac{1}{\varepsilon}}{d_z+1+\frac{1}{\varepsilon}}
        = 24cv^\frac{1}{\varepsilon} C_z \log T^{\frac{1}{d_z+1+\frac{1}{\varepsilon}}} T^\frac{d_z +\frac{1}{\varepsilon}}{d_z+1+\frac{1}{\varepsilon}}.
    \end{align*}
    
    We will consider the last batch $R_M$ separately and make a decision based on this to determine when to proceed to \textbf{Clean-up Phase}. We produce at most $T$ steps in the last batch, so the total regret at $M$-th batch is bounded by
    \begin{align*}
        R_M & \le 8 r_{M-1} T = 8 2^{-\sum_{i=1}^{M-1} c_i} T
        = 8 2^{-c_1 \left(\frac{1-\eta^{M-1}}{1-\eta}\right)} T\\
        & = 8 \left( \frac{T}{\log T} \right)^{ -\frac{1}{d_z+1+\frac{1}{\varepsilon}}(1-\eta^{M-1}) } T \\
        & = 8 \left( \frac{T}{\log T} \right)^{ \frac{\eta^{M-1}}{d_z+1+\frac{1}{\varepsilon}} } (\log T)^{\frac{1}{d_z+1+\frac{1}{\varepsilon}}} T^{\frac{d_z+\frac{1}{\varepsilon}}{d_z+1+\frac{1}{\varepsilon}}}.
    \end{align*}

    Adding regret in each batch, we obtain
    \begin{align*}
        R_T(\pi) = \sum_{m=1}^{M-1} R_m +R_M
        \le 8 \left( 3cv^\frac{1}{\varepsilon} C_z \cdot M +  \left( \frac{T}{\log T} \right)^{ \frac{\eta^{M-1}}{d_z+1+\frac{1}{\varepsilon}} } \right) (\log T)^{\frac{1}{d_z+1+\frac{1}{\varepsilon}}} T^{\frac{d_z+\frac{1}{\varepsilon}}{d_z+1+\frac{1}{\varepsilon}}}.
    \end{align*}
    Above formula can be minimized of $M = \mathcal{O} \left(\log \log \frac{T}{\log T}\right)$. Here we choose $M^* = \frac{\log \log \frac{T}{\log T}}{\log \frac{1}{\eta}}+1 = \frac{\log \log \frac{T}{\log T}}{\log \frac{d+1+\frac{1}{\varepsilon}}{d+1-d_z}}+1$, then $\left( \frac{T}{\log T} \right)^{ \frac{\eta^{M^*-1}}{d_z+1+\frac{1}{\varepsilon}} } = 2^{\frac{1}{d_z+1+\frac{1}{\varepsilon}}}$. With $M^*$, regret is
    \begin{align*}
        R_T(\pi) \le
        \kappa v^\frac{1}{\varepsilon} C_z \log \log \frac{T}{\log T} (\log T)^{\frac{1}{d_z+1+\frac{1}{\varepsilon}}} T^{\frac{d_z+\frac{1}{\varepsilon}}{d_z+1+\frac{1}{\varepsilon}}},
    \end{align*}
    where $\kappa$ is a constant independent of $T$. Constant $\kappa$ in independent of any parameter except constant $c$ about robust mean estimator in Lemma \ref{heavy-tail:property}.
    Consequently, we have
    \begin{align*}
        \mathbb{E} [R_T(\pi)] 
        & = \mathbb{E} [R_T(\pi) | \mathcal{E}] \mathbb{P}(\mathcal{E}) + \mathbb{E} [R_T(\pi) | \mathcal{E}^c] \mathbb{P}(\mathcal{E}^c) \\
        & \le \kappa v^\frac{1}{\varepsilon} C_z \log \log \frac{T}{\log T} (\log T)^{\frac{1}{d_z+1+\frac{1}{\varepsilon}}} T^{\frac{d_z+\frac{1}{\varepsilon}}{d_z+1+\frac{1}{\varepsilon}}}.
    \end{align*}
\end{proof}
 
\section{Lower bound analysis for the Lipschitz setting}
\label{app:lips-lower}

For the lower bound analysis, we focus on the case where $d_z = d$. One can easily extend the argument to $d_z < d$ via the ``decaying'' trick in \cite{10239433}. 

\subsection{static grid}

\begin{proof}[Proof of the static case in Theorem \ref{thm:lower-lips}]

To proof Theorem \ref{thm:lower-lips}, we first show that for any fixed $k\ge1$, there exists an instance such that $\mathbb{E}[R_T(\pi)]\ge \kappa \frac{t_k}{t_{k-1}^{\frac{1}{d_z+1+\frac{1}{\varepsilon}}}}$. 

For each $k$, we construct a sequence of problem instances as follows: Let $r_k=\frac{1}{t_{k-1}^{\frac{1}{d_z+1+\frac{1}{\varepsilon}}}}$ and $M_k:=r_k^{\frac{\varepsilon+1}{\varepsilon}}t_{k-1}=r_k^{-d_z}$. By definition of the dimension $d_z$, we can find a set of arms $\mathcal{U}_k=\{u_{k,1},u_{k,2},\cdots,u_{k,M_k}\}$ such that $d_\mathcal{X}(u_{k,i},u_{k,j})\ge r_k$ for any $i\neq j$. Then we consider a set of problem instance $\mathcal{I}_k=\{I_{k,1},I_{k,2},\cdots,I_{k,M_k}\}$, where the expected reward of $I_{k,1}$ is defined as :
\begin{equation}\label{eq:1}
\mu_{k,1}(x) = 
\begin{cases}
\frac{3}{4}r_k, & \text{if } x = u_{k,1} \\
\frac{5}{8}r_k, & \text{if } x = u_{k,j},\ j \neq 1 \\
\max\left\{ \frac{r_k}{2},\ \max_{u \in \mathcal{U}_k} \{\mu_{k,1}(u) - d_{\mathcal{X}}(x,u)\} \right\}, & \text{otherwise}
\end{cases}
\end{equation}
For $2\le i\le M_k$ the expected reward of $I_{k,i}$ is defined as:
\begin{equation}\label{eq:2}
\mu_{k,i}(x) = 
\begin{cases}
\frac{3}{4}r_k,      &\text{if } x = u_{k,1} \\
\frac{7}{8}r_k,      &\text{if } x = u_{k,i} \\
\frac{5}{8}r_k,      &\text{if } x = u_{k,j},\ j \neq 1 \text{ and } j \neq i \\
\max\left\{ \frac{r_k}{2}, \max_{u \in \mathcal{U}_k} \left\{ \mu_{k,i}(u) - d_{\mathcal{X}}(x,u) \right\} \right\}, & \text{otherwise}
\end{cases}
\end{equation}

For each instance $I_{k,i},1\le i\le M_k$, and any $x\in \mathcal{X}$, the reward distribution is defined as:
\begin{align*}
    \nu_{k,i}(x)=(1-\gamma^{1+\varepsilon}+(2\mu_{k,i}(u_{k,1})-\mu_{k,i}(x))\gamma )\delta_0+(\gamma^{1+\varepsilon}-(2\mu_{k,i}(u_{k,1})-\mu_{k,i}(x))\gamma)\delta_{\frac{1}{\gamma}},
\end{align*}
where $\gamma=(2\mu_{k,i}(u_{k,1}))^\frac{1}{\varepsilon}=(\frac{3}{2}r_k)^\frac{1}{\varepsilon}$ is independent of $i$.

The lower bound is relied on the following lemma.
\begin{lemma}
    For any policy $\pi$, there exist a problem instance $I\in\mathcal{I}_k$ such that
    \begin{align*}
        \mathbb{E}[R_T(\pi)]\ge \frac{r_k}{32}\cdot\sum_{j=1}^M(t_j-t_{j-1})\exp\left(-\frac{1}{4}\cdot\left(\frac{3}{2}\right)^{\frac{1}{\varepsilon}}\frac{r_k^{\frac{\varepsilon+1}{\varepsilon}}t_{j-1}}{M_k-1}\right).
    \end{align*}
\end{lemma}

\begin{proof}
Let $S_{k,i}=\mathbf{B}(u_{k,i},\frac{3}{8}r_k)$ (Ball with center $u_{k,i}$ and radius $\frac{3}{8}r_k$). We can easily verify the following properties by our constructions of problem instances in (\ref{eq:1}),(\ref{eq:2}):
\begin{item}
    1.For any $2\le i\le M_k$, $\mu_{k,i}(x)=\mu_{k,1}(x)$, for any $x\in \mathcal{X}\setminus S_{k,i};$

    2.For any $2\le i\le M_k$, $\mu_{k,1}(x)\le\mu_{k,i}(x)\le\mu_{k,1}(x)+\frac{r_k}{4}$, for any $x\in S_{k,i}$;

    3.For any $1\le i\le M_k$, under $I_{k,i}$, pulling an arm that is not in $S_{k,i}$ incurs a regret at least $\frac{r_k}{8}$, and $\mu_{k,i}(x)\ge \frac{r_k}{2}$, for any $x\in\mathcal{X}$.
\end{item}

Let $\mathbb{P}_{k,i}^t$ denotes the distribution of observations available at time $t$ under instance $I_{k,i}$ and policy $\pi$. And we define $x_t$ as the arm policy $\pi$ choose at time $t$, $y_t$ as the reward, $R^t(\pi)$ as the instantaneous regret incurred by the policy $\pi$ at time $t$. It holds that:
\begin{align}
    \sup_{I\in \mathcal{I}_k}\mathbb{E}[R_T(\pi)]&\ge\frac{1}{M_k}\sum_{i=1}^{M_k}\sum _{t=1}^T\mathbb{E}_{\mathbb{P}_{k,i}^t}[R^t(\pi)]\\
    &\ge\frac{r_k}{8}\sum _{t=1}^T\frac{1}{M_k}\sum_{i=1}^{M_k}\mathbb{P}_{k,i}^t(x_t\notin S_{k,i}).
    \label{eq:7}
\end{align}

From our constructions, note that $S_{k,i}\cap S_{k,j}=\emptyset$, we could easily set a test $\Psi$, such that $x_t\in S_{k,i}$ implies $\Psi=i$. Then according to lemma 3 in \cite{gao2019batched}, we have:
\begin{align*}
    \frac{1}{M_k}\sum_{i=1}^{M_k}\mathbb{P}_{k,i}^t(x_t\notin S_{k,i})\ge \frac{1}{M_k}\sum_{i=1}^{M_k}\mathbb{P}_{k,i}^t(\Psi\neq i)\ge\frac{1}{2M_k}\sum_{i=2}^{M_k}\exp(-D_{KL}(\mathbb{P}_{k,1}^t||\mathbb{P}_{k,i}^t)).
\end{align*}

Now we evaluate $D_{KL}(\mathbb{P}_{k,1}^t||\mathbb{P}_{k,i}^t)$ from the chain rule of KL-Divergence. We have,
\begin{align}
    D_{KL}(\mathbb{P}_{k,1}^t||\mathbb{P}_{k,i}^t)&=D_{KL}\left(\mathbb{P}_{k,1}^{t}(x_1,y_1,\dots,x_{t_{j-1}},y_{t_{j-1}}) \big\| \mathbb{P}_{k,i}^t(x_1,y_1,\dots,x_{t_{j-1}},y_{t_{j-1}})\right) \nonumber \\
    &\le D_{KL}\left(\mathbb{P}_{k,1}^t(x_1,y_1,\cdots,y_{t_{j-1}-1}) \big\| \mathbb{P}_{k,i}^t(x_1,y_1,\cdots,y_{t_{j-1}-1}) \right) \nonumber \\
    &\quad + \mathbb{E}_{\mathbb{P}_{k,1}}\left[D_{KL}\left(\mathbb{P}_{k,1}^t(y_{t_{j-1}}|x_{t_{j-1}}) \big\| \mathbb{P}_{k,i}^t(y_{t_{j-1}}|x_{t_{j-1}})\right)\right] \\
    &\leq D_{KL}\left(\mathbb{P}_{k,1}^t(x_1,y_1,\cdots,y_{j_{t-1}-1}) \big\| \mathbb{P}_{k,i}^t(x_1,y_1,\cdots,y_{j_{t-1}-1})\right) \nonumber \\
    &\quad + \mathbb{E}_{\mathbb{P}_{k,1}}\left[D_{KL}\left(\nu_{k,1}(x_{t_{j-1}})\big\| \nu_{k,i}(x_{t_{j-1}})\right)\right]\\
    &\leq D_{KL}\left(\mathbb{P}_{k,1}^t(x_1,y_1,\cdots,y_{j_{t-1}-1}) \big\| \mathbb{P}_{k,i}^t (x_1,y_1,\cdots,y_{j_{t-1}-1})\right) \nonumber \\
    &\quad + \mathbb{E}_{\mathbb{P}_{k,1}}\left[I_{\{x_{t_{j-1}}\in S_{k,i}\}}D_{KL}\left(\nu_{k,1}(x_{t_{j-1}})\big\| \nu_{k,i}(x_{t_{j-1}})\right)\right]\\
    &\leq D_{KL}\left(\mathbb{P}_{k,1}^t(x_1,y_1,\cdots,y_{j_{t-1}-1}) \big\| \mathbb{P}_{k,i}^t(x_1,y_1,\cdots,y_{j_{t-1}-1})\right) \nonumber \\
    &\quad + \mathbb{P}_{k,1}(x_{t_{j-1}}\in S_{k,i})\cdot\frac{[(\mu_{k,i}(x)-\mu_{k,1}(x))\gamma]^2}{\mu_{k,i}(x)\gamma(1-\mu_{k,i}(x)\gamma)}\\
    &\leq D_{KL}\left(\mathbb{P}_{k,1}^t(x_1,y_1,\cdots,y_{j_{t-1}-1}) \big\| \mathbb{P}_{k,i}^t(x_1,y_1,\cdots,y_{j_{t-1}-1})\right) \nonumber \\
    &\quad + \mathbb{P}_{k,1}(x_{t_{j-1}}\in S_{k,i})\cdot\frac{(\frac{r_k}{4}\gamma)^2}{\frac{r_k}{2}\gamma\cdot\frac{1}{2}}\\
    &\leq D_{KL}\left(\mathbb{P}_{k,1}^t(x_1,y_1,\cdots,y_{j_{t-1}-1}) \big\| \mathbb{P}_{k,i}^t (x_1,y_1,\cdots,y_{j_{t-1}-1})\right) \nonumber \\\label{eq:4}
    &\quad + \frac{1}{4}\cdot\left(\frac{3}{2}\right)^{\frac{1}{\varepsilon}}\cdot r_k^{\frac{\varepsilon+1}{\varepsilon}}\cdot \mathbb{P}_{k,1}(x_{t_{j-1}}\in S_{k,i}).
\end{align}

From formula \ref{eq:4}, we could decompose the KL-Divergence step by step and conclude that:
\begin{align*}
    D_{KL}(\mathbb{P}_{k,1}^t||\mathbb{P}_{k,i}^t)\le\frac{1}{4}\cdot\left(\frac{3}{2}\right)^{\frac{1}{\varepsilon}}\cdot r_k^{\frac{\varepsilon+1}{\varepsilon}}\mathbb{E}_{\mathbb{P}_{k,1}}[\tau_i],
\end{align*}
where $\tau _i$ denotes the number of pulls of arm in $S_{k,i}$ before the batch containing $t$. Then for all $t\in(t_{j-1},t_j]$, we set $\beta=\frac{1}{4}\cdot\left(\frac{3}{2}\right)^{\frac{1}{\varepsilon}}$ and have
\begin{align}
\frac{1}{M_k} \sum_{i=1}^{M_k} \mathbb{P}_{k,t}^t(x_t \notin S_{k,i}) 
&\geq \frac{1}{2M_k} \sum_{i=2}^{M_k} \exp\left(-\beta r_k^{\frac{\varepsilon+1}{\varepsilon}} \mathbb{E}_{\mathbb{P}_{k,1}}[\tau_1]\right) \nonumber \\
&\geq \frac{M_k -1}{2M_k} \exp\left(-\frac{\beta r_k^{\frac{\varepsilon+1}{\varepsilon}}}{(M_k -1)} \sum_{i=2}^{M_k} \mathbb{E}_{\mathbb{P}_{k,1}}[\tau_i]\right)\\
&\geq \frac{1}{4} \exp\left(-\frac{\beta r_k^{\frac{\varepsilon+1}{\varepsilon}}(t_j -1)}{(M_k -1)}\right) .\label{eq:6}
\end{align}
We substitute \ref{eq:6} to \ref{eq:7} to finish the proof of lemma.
\end{proof}

Since $M_k=r_k^{\frac{\varepsilon+1}{\varepsilon}}t_{k-1}=r_k^{-d_z}$, the result of lemma leads to $\mathbb{E}[R_T(\pi)]\ge \kappa \varepsilon^{\frac{1}{2}} \frac{t_k}{t_{k-1}^{\frac{1}{d_z+1+\frac{1}{\varepsilon}}}}$ directly.
Then, as it could be applied to any $k$, we could find an instance such that
\begin{align*}
    \mathbb{E}[R_T(\pi)]\ge \kappa \varepsilon^{\frac{1}{2}} \max_{1\le k\le M}\left\{\frac{t_k}{t_{k-1}^{\frac{1}{d_z+1+\frac{1}{\varepsilon}}}}\right\},
\end{align*}
which will lead to the result of Theorem \ref{thm:lower-lips}.
\end{proof}

\subsection{adaptive grid}

\begin{proof}[Proof of adaptive case in Theorem \ref{thm:lower-lips}]

For adaptive grid $\mathcal{T}=\{ t_0,t_1,\cdots,t_M \}$, we consider a reference static grid $\mathcal{T}_r=\{T_0,T_1,\cdots,T_M \}$, where $T_j=T^{\frac{1-\left( \frac{1}{d_z+1+\frac{1}{\varepsilon}} \right)^j}{1-\left( \frac{1}{d_z+1+\frac{1}{\varepsilon}} \right)^M}}$. We define $a=\frac{1}{d_z+1+\frac{1}{\varepsilon}}$, so that $T_j=T^{\frac{1-a^j}{1-a^M}}$. Further we define $r_j = \frac{1}{T_{j-1}^a M}$ and $M_j = \frac{1}{r_j^{d_z}}$. We get
\begin{align*}
    T_{j-1} r_j^\frac{1+\varepsilon}{\varepsilon} = r_j^{-\frac{1}{a}+\frac{1+\varepsilon}{\varepsilon}} M^{-\frac{1}{a}} = \frac{M_j}{M^\frac{1}{a}}.
\end{align*}

For $1\le j \le M$, we construct a series of instance $\mathcal{I}_j$ and investigate the event $A_j = \{ t_{j-1} < T_{j-1}, t_j \ge T_j \}$. First, let's introduce the construction of $\mathcal{I}_1, \cdots, \mathcal{I}_M$.

For $1 \le j \le M$, we can find arms $\mathcal{U}_j=\left\{u_{j, 1}, \cdots, u_{j, M_j}\right\}$ such that $d_{\mathcal{A}}\left(u_{j, m}, u_{j, n}\right) \geq r_j$ for any $m \neq n$, and $u_{1, M_1}=\cdots=u_{M, M_M}$. For $1 \leq j \leq M-1$, $\mathcal{I}_j$ has $M_j-1$ instances $\mathcal{I}_j=\left\{I_{j, k}\right\}_{k=1}^{M_j-1}$, and the expected reward of $I_{j, k}$ is defined as
\begin{align*}
    \mu_{j, k}(x)= \begin{cases}\frac{r_1}{2}+\frac{r_j}{16}+\frac{r_M}{16}, & \text{if } x=u_{j, k} \\ \frac{r_1}{2}+\frac{r_M}{16}, & \text{if } x=u_{j, M_j} \\
    \max \left\{\frac{r_1}{2}, \max _{u \in \mathcal{U}_j} \left\{\mu_{j, k}(u)-d_{\mathcal{X}}(x, u)\right\}\right\}, & \text{otherwise. }
    \end{cases}    
\end{align*}
For $j=M$, we let $\mathcal{I}_M=\left\{I_{j,M}\right\}_{j=1}^{M-1}$ and the expected reward of $I_M$ is defined as
\begin{align*}
    \mu_M\left(x\right)=
    \begin{cases}
        \frac{r_1}{2} + \frac{r_M}{16}, & \text{if } x=u_{M, M_M} \\
        \max \left\{\frac{r_1}{2}, \mu_M\left(u_{M, M_M}\right)-d_{\mathcal{X}}\left(x, u_{B, M_B}\right)\right\},
        & \text{otherwise.}
    \end{cases}
\end{align*}
Distribution of each reward is heavy-tailed. We take the following form.
For instance $I_{j,k}, 1\le j \le M-1, 1\le k\le M_j-1$, at any $x\in \mathcal{X}$ the reward distribution is defined as:
\begin{align*}
    \nu_{j,k}(x)& =(1-\mu_{j,k}(x)\gamma_j -\mu_M(x)\sum_{i\neq j}\gamma_i )\delta_0+(\mu_{j,k}(x)\gamma_j)\delta_{\frac{1}{\gamma_j}}
    + \sum_{i \neq j} ( \mu_M(x)\gamma_i ) \delta_{\frac{1}{\gamma_i}} \\
    &\quad- (M-1) \mu_M(x),
\end{align*}
where $\gamma_j=r_j^{\frac{1}{\varepsilon}} , j=1,\cdots,M$ is independent of $k$.
For instance $I_M$, we construct the reward distribution at any $x\in \mathcal{X}$ as:
\begin{align*}
    \nu_{M}(x)=(1-\mu_{M}(x) \sum_{j=1}^M \gamma_j )\delta_0+ \sum_{j=1}^M (\mu_{M}(x)\gamma_j)\delta_{\frac{1}{\gamma_j}} -(M-1)\mu_M(x).
\end{align*}
Above reward distributions satisfy the required conditions while remaining computationally tractable. Expectation of $\nu_{j,k}(x)$ (resp. $\nu_M(x)$) is $\mu_{j,k}(x)$ (resp. $\mu_M(x)$), and they have bounded $1+\varepsilon$ moments and unbounded variance.
Through techniques such as Taylor expansion and straightforward calculations, we can obtain 
\begin{align*}
    D_{KL}(\nu_{j,k}(x)||\nu_M(x)) \le \kappa \frac{(\mu_{j,k}(x)-\mu_M(x))^2\gamma_j}{\mu_M(x)},
\end{align*}
where $\kappa$ is a numerical constant.

Similar to the previous text, these constructions of $\mu$ satisfy two properties: 
1. $\mu_{j, k}$ is close to $\mu_M$;
2. under $I_{j, k}$, pulling an arm that is far from $u_{j, k}$ incurs a regret at least $\frac{r_j}{16}$.

We then show that for any adaptive grid $\mathcal{T}=\left\{t_0, \cdots, t_M\right\}$, there exists an index $j$ such that $\left(t_{j-1}, t_j\right]$ is sufficiently large in $\mathcal{I}_j$. Specifically, we consider event $A_j=\left\{t_{j-1}<T_{j-1}, t_j \geq T_j\right\}$. To evaluate $A_j$, we define $p_j:=\frac{1}{M_j-1} \sum_{k=1}^{M_j-1} \mathbb{P}_{j, k}\left(A_j\right)$ for $j \le M-1$ and $p_M:=\mathbb{P}_M\left(A_M\right)$. We have the following lemma to illustrate the probability of $p_j$.

\begin{lemma}
    For any adaptive grid $\mathcal{T}=\left\{t_0, \cdots, t_M\right\}$ and policy $\pi$, it holds that $\sum_{i=j}^M p_j \ge \frac{7}{8}$.
\end{lemma}

\begin{proof}
Similar as before, we define the ball $S_{j, k}:=\mathbb{B}\left(u_{j, k}, \frac{3}{8} r_j\right)$. It is easy to verify that outside $S_{j, k}$, $\mu_{j, k}=\mu_M$. Whereas for any $x \in S_{j, k}$, $\mu_M(x) \leq \mu_{j, k}(x) \leq \mu_M(x)+\frac{r_j}{8}$.

The notation of $x_t$ and $y_t$ denote the same meaning as mentioned before. For $t_{j-1}<t \leq t_j$, we define $\mathbb{P}_{j, k}^t$ (resp. $\mathbb{P}_M^t$ ) as the distribution of sequence ( $x_1, y_1, \cdots, x_{t_{j-1}}, y_{t_{j-1}}$ ) under instance $I_{j, k}$ (resp. $I_M$ ) and policy $\pi$. We compare event $A_j$ under $\mathbb{P}_{j,k}$ and $\mathbb{P}_M$ to get
\begin{align}
    |\mathbb{P}_M(A_j) - p_j| \le 
    \frac{1}{M_j-1} \sum_{k=1}^{M_j-1} |\mathbb{P}_M(A_j) - \mathbb{P}_{j,k}(A_j)|.
    \label{eq:lips-lower-ada-1}
\end{align}
Since data is obtained in batches, event $A_j$ can be completely described by the observations up to time $T_{j-1}$.
Using the total variation, we have
\begin{align*}
    \left|\mathbb{P}_M\left(A_j\right)-\mathbb{P}_{j, k}\left(A_j\right)\right|
    =\left|\mathbb{P}_M^{T_{j-1}}\left(A_j\right)-\mathbb{P}_{j, k}^{T_{j-1}}\left(A_j\right)\right| \leq T V\left(\mathbb{P}_M^{T_{j-1}}, \mathbb{P}_{j, k}^{T_{j-1}}\right).
\end{align*}

Combine Eq. (\ref{eq:lips-lower-ada-1}), we get
\begin{align*}
    \frac{1}{M_j-1} \sum_{k=1}^{M_j-1} |\mathbb{P}_M(A_j) - \mathbb{P}_{j,k}(A_j)|
    & \le
    \frac{1}{M_j-1} \sum_{k=1}^{M_j-1} T V\left(\mathbb{P}_M^{T_{j-1}}, \mathbb{P}_{j, k}^{T_{j-1}}\right) \\
    &\le 
    \frac{1}{M_j-1} \sum_{k=1}^{M_j-1} \sqrt{1-\exp \left(-D_{K L}\left(\mathbb{P}_M^{T_{j-1}} \| \mathbb{P}_{j, k}^{T_{j-1}}\right)\right)}.
\end{align*}

An argument similar to Eq. (\ref{eq:4}) yields that in $S_{j,k}$:
\begin{align*}
    D_{K L}\left(\mathbb{P}_M^{T_{j-1}} \| \mathbb{P}_{j, k}^{T_{j-1}}\right) \le \kappa r_j^{\frac{1+\varepsilon}{\varepsilon}}\mathbb{E}_{\mathbb{P}_M \tau_k},
\end{align*}
where $\tau_k$ denotes the number of pulls which is in $S_{j, k}$ before the batch containing $T_{j-1}$ and $\kappa$ is a numerical constant. Such that, we get
\begin{align*}
    |\mathbb{P}_M(A_j) - p_j|
    & \le
    \frac{1}{M_j-1} \sum_{k=1}^{M_j-1} \sqrt{1-\exp \left(- \kappa r_j^{\frac{1+\varepsilon}{\varepsilon}}\mathbb{E}_{\mathbb{P}_M \tau_k} \right)} \\
    & \le
    \sqrt{1-\exp \left(-\frac{ \kappa r_j^{\frac{1+\varepsilon}{\varepsilon}}}{M_j-1} \sum_{k=1}^{M_j-1} \mathbb{E}_{\mathbb{P}_M \tau_k} \right)}\\
    & \le
    \sqrt{1-\exp \left(-\frac{ \kappa r_j^{\frac{1+\varepsilon}{\varepsilon}}}{M_j-1} T_{j-1}\right)} \\
    & \le
    \sqrt{1-\exp \left(- \kappa M^{-\frac{1}{a}}\right)}
    \le \frac{1}{8M}.
\end{align*}

Finally we can use $\sum_{j=1}^M\mathbb{P}_M(A_j)\ge1$ to directly obtain 
\begin{align*}
    \sum_{j=1}^M p_j \ge \sum_{j=1}^M\mathbb{P}_M(A_j) - \sum_{j=1}^M(\mathbb{P}_M(A_j) - p_j) 
    \ge \frac{7}{8}.
\end{align*}

\end{proof}

Then we turn to the lower bound we need.

For any $1 \leq k \leq M_j-1$, we construct a set of instances $\mathcal{I}_{j, k}=\left(I_{j, k, l}\right)_{1 \leq l \leq M_j}$ according to $\mathcal{I}_j$. For $l \neq k$, the expected reward of $I_{j, k, l}$ is defined as
\begin{align*}
    \mu_{j, k, l}(x)=
    \begin{cases}
\mu_{j, k}(x)+\frac{3 r_j}{16}, &\text { if } x=u_{j, l} \\
\mu_{j, k}(x), &\text { if } x \in \mathcal{U}_j \text { and } x \neq u_{j, l} \\
\max \left\{\frac{r_1}{2}, \max _{u \in \mathcal{U}_j}\left\{\mu_{j, k, l}(u)-d_{\mathcal{X}}(x, u)\right\}\right\}, &\text { otherwise, }
\end{cases}
\end{align*}
where $\mu_{j, k}$ is defined ahead. For $l=k$, we set $\mu_{j, k, k}=\mu_{j, k}$.

We define the ball $C_{j, k}=\mathbb{B}\left(u_{j, k}, \frac{r_j}{4}\right)$. Then in $\mathcal{I}_{j, k}$ we have that for any $l \neq k, \mu_{j, k, l}(x)=\mu_{j, k, k}(x)$ when $x \notin C_{j, l}$.
Also, $\mu_{j, k, k}(x) \leq \mu_{j, k, l}(x) \leq \mu_{j, k, k}(x)+\frac{3 r_j}{16}$ in $C_{j, l}$, which implies pulling an arm that is not in $C_{j, l}$ incurs a regret at least $\frac{r_j}{16}$.
Analogously, for $t_{j-1}<t \leq t_j$, we define $\mathbb{P}_{j, k, l}^t$ as the distribution of sequence $\left(x_1, y_1, \cdots, x_{t_{j-1}}, y_{t_{j-1}}\right)$ under instance $I_{j, k, l}$ and policy $\pi$.

From similar argument, it holds that
\begin{align*}
    \sup _{I \in \mathcal{I}_{j, k}} \mathbb{E}\left[R_T(\pi)\right] \geq \frac{r_j}{16} \sum_{t=1}^T \frac{1}{M_j} \sum_{l=1}^{M_j} \mathbb{P}_{j, k, l}^t\left(x_t \notin C_{j, l}\right) .
\end{align*}

By Lemma in \cite{gao2019batched} and $\int \min \{ dP, dQ \} = 1-TV(P,Q)$, we have
\begin{align*}
    \frac{1}{M_j} \sum_{l=1}^{M_j} \mathbb{P}_{j, k, l}^t\left(x_t \notin C_{j, l}\right) 
    & \ge \frac{1}{M_j} \sum_{l \neq k} \int \min \left\{d \mathbb{P}_{j, k, k}^t, d \mathbb{P}_{j, k, l}^t\right\} \\
    & \ge
    \frac{r_j T_j}{16 M_j} \sum_{j \neq k} \int \min \left\{d \mathbb{P}_{j, k, k}^{T_j}, d \mathbb{P}_{j, k, l}^{T_j}\right\} \\
    & \ge \frac{r_j T_j}{16 M_j} \sum_{j \neq k} \int_{A_j} \min \left\{d \mathbb{P}_{j, k, k}^{T_j}, d \mathbb{P}_{j, k, l}^{T_j}\right\}.
\end{align*}

To estimate above formula, first we denote definition of total variance to obtain
\begin{align*}
    \int_{A_j} \min \left\{d \mathbb{P}_{j, k, k}^{T_j}, d \mathbb{P}_{j, k, l}^{T_j}\right\}
    & = \frac{\mathbb{P}_{j, k, k}^{T_j}(A_j) + \mathbb{P}_{j, k, l}^{T_j}(A_j)}{2} - \frac{1}{2} \int_{A_j}  \left| d \mathbb{P}_{j, k, k}^{T_j} - d \mathbb{P}_{j, k, l}^{T_j}\right| \\
    & \ge 
    \mathbb{P}_{j, k}(A_j) -\frac{3}{2} TV(\mathbb{P}_{j, k, k}^{T_j}, d \mathbb{P}_{j, k, l}^{T_j}).
\end{align*}
And by employing a distribution construction similar to the previous approach, we can achieve
\begin{align*}
    D_{K L}\left(\mathbb{P}_{j, k, k}^{T_{j-1}} \| \mathbb{P}_{j, k, l}^{T_{j-1}}\right) 
    \le \kappa r_j^{\frac{1+\varepsilon}{\varepsilon}} \mathbb{E}_{\mathbb{P}_{j, k}} \tau_l,
\end{align*}
where $\tau_l$ denotes the number of pulls which is in $C_{j, l}$ before the batch of time $T_{j-1}$ and $\kappa$ is a numerical constant. That is to say,
\begin{align*}
    \frac{1}{M_j} \sum_{l \neq k} T V\left(\mathbb{P}_{j, k, k}^{T_{j-1}}, \mathbb{P}_{j, k, l}^{T_{j-1}}\right) 
    & \le \frac{1}{M_j} \sum_{l \neq k} \sqrt{1-\exp \left(-D_{K L}\left(\mathbb{P}_{j, k, k}^{T_{j-1}} \| \mathbb{P}_{j, k, l}^{T_{j-1}}\right)\right)} \\
    & \le \frac{1}{M_j} \sum_{l \neq k} \sqrt{1-\exp \left(- \kappa r_j^{\frac{1+\varepsilon}{\varepsilon}} \mathbb{E}_{P_{j, k}} \tau_l\right)} \\
    & \le \frac{M_j-1}{M_j} \sqrt{1-\exp \left(-\frac{ \kappa r_j^{\frac{1+\varepsilon}{\varepsilon}} T_{j-1}}{\left(M_j-1\right)}\right)} \\
    & \le \sqrt{1-\exp \left(- \kappa M^{-(d_z+1+\frac{1}{\varepsilon})}\right)} \le \frac{1}{4 M}.
\end{align*}

Building upon the preliminary results established earlier,
\begin{align*}
    \sup _{I \in \mathcal{I}_{j, k}} \mathbb{E}\left[R_T(\pi)\right] 
    \ge \frac{1}{16} r_j T_j\left(\mathbb{P}_{j, k}\left(A_j\right)-\frac{3}{8 M}\right).
\end{align*}
Thanks to our definition, $r_j T_j = \frac{1}{M} T^{\frac{1-a}{1-a^M}}$, where $a=\frac{1}{d_z+1+\frac{1}{\varepsilon}}$.Above inequality holds for any $k \leq M_j-1$. Here we average over $k$ and show
\begin{align*}
    \sup _{I \in \cup_{k \leq M M_j-1} \mathcal{I}_{j, k}} \mathbb{E}\left[R_T(\pi)\right] 
    & \ge \frac{1}{M_j-1} \sum_{k=1}^{M_j-1} \sup _{I \in \mathcal{I}_{j, k}} \mathbb{E}\left[R_T(\pi)\right]\\
    & \ge \frac{1}{16 M} T^{\frac{1-a}{1-a^M}} \left(\frac{1}{\left(M_j-1\right)} \sum_{k=1}^{M_j-1} \mathbb{P}_{j, k}\left(A_j\right)-\frac{3}{8 M}\right) \\ 
    & \ge \kappa \frac{1}{M^2} T^{\frac{1-\frac{1}{d_z+1+\frac{1}{\varepsilon}}}{1-\left( \frac{1}{d_z+1+\frac{1}{\varepsilon}}\right)^M}}.
\end{align*}
Thus, the proof is complete.

\end{proof}

\subsection{Proof of Corollary \ref{cor:lower-lips-M}}

\begin{proof}
    We hope to find requirements for $M$ such that 
    \begin{align*}
        T^{\frac{1-a}{1-a^M}} \le C T^{\frac{1-a}{1-a^T}},
    \end{align*}
    for come constant $C$. Above $a = \frac{1}{d_z+1+\frac{1}{\varepsilon}}$ is a constant within $[0,1]$.
    The preceding expression admits the following equivalent representation:
    \begin{align*}
        \frac{1-a}{1-a^M} \le \frac{\log C}{\log T} +\frac{1-a}{1-a^T}.
    \end{align*}
    Equivalently, the expression can be 
    \begin{align*}
        a^M \le
        \frac{\frac{\log C}{\log T} +\frac{1-a}{1-a^T}-(1-a)}{\frac{\log C}{\log T} +\frac{1-a}{1-a^T}}.
    \end{align*}
    Then we take the logarithm on both sides. Since $a$ is below to $1$, lower bound of number of batches satisfies
    \begin{align*}
        M \ge 
        \log \frac{1}{a} \cdot \log \frac{\frac{\log C}{\log T} +\frac{1-a}{1-a^T}}{\frac{\log C}{\log T} +\frac{1-a}{1-a^T}a^T}
        = 
        \log^{-1} \frac{1}{a} \cdot \log \frac{\log C +\frac{1-a}{1-a^T} \log T}{\log C +\frac{1-a}{1-a^T}a^T \log T},
    \end{align*}
    for come constant $C$.
    When $T$ is large enough, $a^T \log T = o(1)$. Bring $a=\frac{1}{d_z+1+\frac{1}{\varepsilon}}$ into the formula, we get
    \begin{align*}
        M \ge C
        \log^{-1} \left( d_z+1+\frac{1}{\varepsilon} \right) \log \log T,
    \end{align*}
    for come constant $C$. Furthermore, we declare the analysis on $T^{\frac{1-a}{1-a^T}}$ and $T^{1-a}$, which is
    \begin{align*}
        \frac{ T^{\frac{1-a}{1-a^T}} }{T^{1-a}}
        = T^{(1-a)\frac{a^T}{1-a^T}} \le C
    \end{align*}
    for some constant $C$.
\end{proof}

\section{Proof of Lemmas}
\label{app:lem}
\subsection{Proof of Lemma \ref{lem:upper}}
\begin{proof}
    We calculate $\mathbb{P} (E)$ by $\mathbb{P} (E^c)$. First we show that $\mathbb{P} \left( B^c \right)$ is small. If the optimal arm $\star$ is eliminated, it must be eliminated by arm $i$ at time $t$,
\begin{align*}
    \mathbb{P}(B^c) = \mathbb{P} ( \text{arm} \star \text{is eliminated})
    \le \sum_{i=1}^K \mathbb{P} ( \text{arm} \star \text{is eliminated by arm } i ). 
\end{align*}

At time $t$ both arms are pulled at least $\tau(t)$ times. Here $t$ is a grid $\in \mathcal{T}$ because we only carry out eliminations after each batch. By our elimination rule:
\begin{align*}
    \mathbb{P}(B^c)
    &\le \sum_{i=1}^K \sum_{t=1}^T \mathbb{P} ( \text{arm} \star \text{is eliminated by arm } i \text{ at time } t) \\
    &\le \sum_{i=1}^K \sum_{t=1}^T \mathbb{P} \left( \hat{\mu}_{i}(\tau(t)) - \hat{\mu}_{\star}(\tau(t)) \ge 2^{\frac{1+2\varepsilon}{1+\varepsilon}} v^{\frac{1}{1+\varepsilon}} \left( \frac{c \log(TK)}{\tau(t)} \right)^{\frac{\varepsilon}{1+\varepsilon}} \right).\\
\end{align*}

For any fixed realization of $\tau(t)$ , each item occurs with probability at most
\begin{align*}
    & \mathbb{P} \left( \hat{\mu}_{i}(\tau(t)) - \hat{\mu}_{\star}(\tau(t)) \ge 2^{\frac{1+2\varepsilon}{1+\varepsilon}} v^{\frac{1}{1+\varepsilon}} \left( \frac{c \log(TK)}{\tau(t)} \right)^{\frac{\varepsilon}{1+\varepsilon}} \right) \\
    = \; & \mathbb{P} \left( \hat{\mu}_{i}(\tau(t)) - \mu_i - \left( \hat{\mu}_{\star}(\tau(t)) -\mu_{\star} \right) - \triangle_i \ge 2 \cdot 2^{\frac{\varepsilon}{1+\varepsilon}} v^{\frac{1}{1+\varepsilon}} \left( \frac{c \log(TK)}{\tau(t)} \right)^{\frac{\varepsilon}{1+\varepsilon}} \right)\\
    \le \; & \mathbb{P} \left( \hat{\mu}_{i}(\tau(t)) - \mu_i \ge 2^{\frac{\varepsilon}{1+\varepsilon}} v^{\frac{1}{1+\varepsilon}} \left( \frac{c \log(TK)}{\tau(t)} \right)^{\frac{\varepsilon}{1+\varepsilon}} \right) \\
    & + 
    \mathbb{P} \left( \hat{\mu}_{\star}(\tau(t)) -\mu_{\star} \le - 2^{\frac{\varepsilon}{1+\varepsilon}} v^{\frac{1}{1+\varepsilon}} \left( \frac{c \log(TK)}{\tau(t)} \right)^{\frac{\varepsilon}{1+\varepsilon}} \right) \\
    \le \; & \left( \frac{1}{TK} \right)^2 + \left( \frac{1}{TK} \right)^2 = 2 \left( \frac{1}{TK} \right)^2.
 \end{align*}
Here we use the properties of best arm and mean estimators.
As a result, 
\begin{equation*}
    \mathbb{P}(B^c) \le TK \cdot 2 \left( \frac{1}{TK} \right)^2 = \frac{2}{TK}.
\end{equation*}

Next we upper bound $\mathbb{P}(B \cup A_i^c)$ separately. With $A_i$'s definition, 
\begin{align*}
    \mathbb{P}(B \cup A_i^c) 
    & = \mathbb{P} \left( \star \text{exists and arm $i$ is not eliminated at time } t_{m_i} \right).
\end{align*}
Here $t_{m_i}$ also is a batched gird. At time $t_{m_i}$, suppose arm $i$ has been pulled $\tau$ times, we have $\tau \ge \tau_i^{\ast}$ and $\tau(t_{m_i}) \ge \tau_i^{\ast}-1 $. We remark that $m_i$ and $\tau(t_{m_i})$ are random variables depending on the random rewards. 
\begin{align*}
    \mathbb{P}(B \cup A_i^c) 
    & = \mathbb{P} \left( \star \text{ exists, } \hat{\mu}_{max}(\tau(t_{m_i})) - \hat{\mu}_i(\tau(t_{m_i})) \le 2^{\frac{1+2\varepsilon}{1+\varepsilon}} v^{\frac{1}{1+\varepsilon}} \left( \frac{c \log(TK)}{\tau(t_{m_i})} \right)^{\frac{\varepsilon}{1+\varepsilon}} \right)\\
    & \le \mathbb{P} \left( \hat{\mu}_{\star}(\tau(t_{m_i})) - \hat{\mu}_i(\tau(t_{m_i})) \le 2^{\frac{1+2\varepsilon}{1+\varepsilon}} v^{\frac{1}{1+\varepsilon}} \left( \frac{c \log(TK)}{\tau(t_{m_i})} \right)^{\frac{\varepsilon}{1+\varepsilon}} \right) \\
    & \le \sum_{\tau = \tau_i^{\ast}-1}^T \mathbb{P} \left( \hat{\mu}_{\star}(\tau) - \hat{\mu}_i(\tau) \le 2^{\frac{1+2\varepsilon}{1+\varepsilon}} v^{\frac{1}{1+\varepsilon}} \left( \frac{c \log(TK)}{\tau} \right)^{\frac{\varepsilon}{1+\varepsilon}} \right).
\end{align*}
Consistent with our earlier proof technique, each item in the above expression is upper bounded by $ 2 \left( \frac{1}{TK} \right)^2$ through a combination of $\triangle_i = 2^{\frac{2+3\varepsilon}{1+\varepsilon}} v^{\frac{1}{1+\varepsilon}} \left( \frac{c \log(TK)}{\tau_i^{\ast}} \right)^{\frac{\varepsilon}{1+\varepsilon}}$ and the properties of mean estimator.
Therefore, by a union bound,
\begin{equation*}
    \mathbb{P}(B \cup A_i^c) \le T \cdot 2 \left( \frac{1}{TK} \right)^2 = \frac{2}{TK^2}.
\end{equation*}

We conclude that
\begin{align*}
    \mathbb{P}(E^c) \le \mathbb{P}(B^c) + \sum_{i=1}^K  \mathbb{P}(B \cap A_i^c ) \le \frac{2}{TK} + K \cdot \frac{2}{TK^2} = \frac{4}{TK},
\end{align*} 
\begin{align*}
    \mathbb{P}(E) = 1-\mathbb{P}(E^c) \ge 1-\frac{4}{TK} .
\end{align*}

When the event E does not occur, the expected regret is at most 
\begin{align*}
    \mathbb{E} \left[R_T\left(\pi\right) \mathbb{I}(E^c) \right] \le T \max_{i \in [K]} \triangle_i \cdot \mathbb{P}(E^c) = o(1).
\end{align*}
Later we always assume that $E$ holds.
\end{proof}

\subsection{Proof of Lemma \ref{lem:lower-static}}

\begin{proof}[Proof of Lemma \ref{lem:lower-static}]
To proof \textbf{Lemma} \ref{lem:lower-static}, we propose distributions $V_1$, $V_2$, and $V_3$ at first. Here, 
\begin{itemize}
    \item[]     $V_1=(1-r^ {1+\varepsilon} )\delta_0+r^ {1+\varepsilon} \delta_{\frac{1}{r}}$,
    \item[]     $V_2=(1-r^ {1+\varepsilon}+\triangle r )\delta_0+(r^ {1+\varepsilon}-\triangle r) \delta_{\frac{1}{r}}$,
    \item[]     $V_3=(1-r^ {1+\varepsilon}+2\triangle r )\delta_0+(r^ {1+\varepsilon}-2\triangle r) \delta_{\frac{1}{r}}$,
\end{itemize}
where $r=(3\triangle)^{\frac{1}{\varepsilon}}$.

Then we estimate $D_{KL}(V_1||V_2)$ and $D_{KL}(V_1||V_3)$. According to the inequality $D_{KL}(Ber(p),Ber(q))\leq \frac{(p-q)^2}{q(1-q)}$, we have
\begin{align*}
    D_{KL}(V_1||V_2)&=\sum_{} p_x \log \frac{p_x}{q_x}\\
    &=(1-r^{1+\varepsilon})\log\left( \frac{1-r^{1+\varepsilon}}{ 1-r^{1+\varepsilon}+\triangle r} \right)+r^{1+\varepsilon}\log\left( \frac{1-r^{1+\varepsilon}}{ 1-r^{1+\varepsilon}+\triangle r} \right),\\
    D_{KL}(V_1||V_3)&\leq \frac{(2\triangle r)^2}{ (r^{1+\varepsilon}-2\triangle r)(1-r^{1+\varepsilon}+2\triangle r) }\\
    &=\frac{4(\triangle r)^2}{\triangle r(1-\triangle r)}\\
    &=\frac{4\triangle r}{1-\triangle r}\\
    &\leq 8\triangle r.
\end{align*}
Consider the following $K$ candidate reward distributions:
\begin{itemize}
    \item[]     $P_1=V_2\otimes V_3\otimes V_3\otimes \cdots \otimes V_3$
    \item[]     $P_2=V_2\otimes V_1\otimes V_3\otimes \cdots \otimes V_3$
    \item[]     $P_3=V_2\otimes V_3\otimes V_1\otimes \cdots \otimes V_3$
    \item[]     $\cdots$
    \item[]     $P_K=V_2\otimes V_3\otimes V_3\otimes \cdots \otimes V_1$
\end{itemize}
The proposed construction possesses two fundamental properties:
\begin{itemize}
    \item[1.] $\forall i\in [K]$,arm $i$ is the optimal arm under reward distribution $P_i$;
    \item[2.] $\forall i\in [K]$, pulling a wrong arm incurs a regret at least $\triangle$ under reward distribution $P_i$.
\end{itemize}
As a result, we have 
\begin{align} 
    \label{eq:static-1} 
 \sup_{\{\mu^{(i)}\}^K_{i=1}:\triangle_i\in\{0\}\cup[\triangle,K^{\varepsilon/(1+\varepsilon)}]}\mathbb{E}R_T(\pi)\geq \frac{1}{ K}\sum^K_{i=1}\sum^T_{t=1}\mathbb{E}_{P^t_i}R^t(\pi)\geq \triangle\sum^T_{t=1}\frac{1}{ K}\sum^K_{i=1}P^t_i(\pi_t\not=i). 
\end{align}
where $P^t_i$ denotes the distribution of observations available at time $t$ under $P_i$, and $R^t(\pi)$ denotes the instantaneous regret incurred by the policy $\pi_t$ at time $t$.
To estimate Eq.(\ref{eq:static-1}), we take star tree $T_r=([K],\{(1,i):2\leq i\leq K\})$.
For $i\in [K]$, denote by $T_i(t)$ the number of pulls of arms $i$ anterior to the current batch of $t$. Hence, $\sum^K_{i=1}T_i(t)=t_{j-1}$ if $t\in(t_{j-1},t_j]$. Thus according to Lemma 3 in \cite{gao2019batched}, we have
\begin{align}
    \begin{aligned}
        \label{eq:static-2}
        \frac{1}{ K} \sum^K_{i=1}P^t_i(\pi_t\not=i)&\geq{\frac{1}{ 2K}}\sum^K_{i=2}\exp(-D_{KL}(P^t_1||P^t_i))\\
    &=\frac{1}{ 2K}\sum^K_{i=2}\exp\left(-D_{KL}(P^t_1||P^t_i)\right)\\
    &\geq \frac{1}{ 2K}\sum^K_{i=2}\exp\left(-8\triangle r\mathbb{E}_{P^t_1}T_i(t)\right)\\
    &=\frac{1}{ 2K}\sum^K_{i=2}\exp\left(-8\cdot3^{\frac{1}{\varepsilon}} \triangle^{\frac{1+\varepsilon}{ \varepsilon}} \mathbb{E}_{P^t_1}T_i(t)\right)\\
    &\geq \frac{K-1}{ 2K}\exp\left(-\frac{C 3^{\frac{1}{\varepsilon}} \triangle^{\frac{1+\varepsilon}{ \varepsilon}}}{ K-1}\sum^K_{i=2}\mathbb{E}_{P^t_1}T_i(t)\right)\\
    &\geq \frac{1}{ 4}\exp\left(- \frac{C 3^{\frac{1}{\varepsilon}} \triangle^{\frac{1+\varepsilon}{ \varepsilon}}}{ K-1}t_{j-1}\right).
    \end{aligned}
\end{align}
Combining Eq.(\ref{eq:static-1}) and Eq.(\ref{eq:static-2}) gives the proof of \textbf{Lemma} \ref{lem:lower-static}.
\end{proof}

\subsection{Proof of Lemma \ref{lem:lower-pro}}
\begin{proof}

 Note that the event $A_{j}$ could be determined by the observations up to time $T_{j-1}$, the data-processing inequality gives
 \begin{align*}|P_{M}(A_{j})-P_{j,k}(A_{j})|\leq{}TV(P_{M}^{T_{j-1}},P_{j,k}^{T_{j-1}}).\end{align*}

 Recall that each $P_{j,k}$ only differs from $P_{M}$ in the k-th component with mean difference $\triangle_{j}+\triangle_
 {M}$, by the same argument in the \textbf{Lemma} \ref{lem:lower-static}, we have
\begin{align}
    \frac{1}{K-1}\sum_{k=1}^{K-1}TV(P_{M}^{T_{j-1}},P_{j,k}^{T_{j-1}})&\leq\frac{1}{K-1}\sum_{k=1}^{K-1}\sqrt{1-\exp\left(-D_{KL}(P_{M}^{T_{j-1}}\|P_{j,k}^{T_{j-1}})\right)}\\
    &\overset{(a)}{\leq}\frac{1}{K-1}\sum_{k=1}^{K-1}\sqrt{1-\exp\left(-\alpha(\triangle_{M}+\triangle_{j})^{\frac{1+\varepsilon}{\varepsilon}}\mathbb{E}_{P_{M}}[\tau_{k}]\right)}\\
    &\overset{(b)}{\leq}\sqrt{1-\exp\left(-\frac{ 2^{\frac{1+\varepsilon}{\varepsilon}}\alpha(\triangle_{j})^{\frac{1+\varepsilon}{\varepsilon}}}{K-1} \mathbb {E_{P_{M}}} \left[ {\sum_{k=1}^{K-1}\tau_{k}} \right ] \right)}\\
    &\leq{} \sqrt{1-\exp\left(- \frac{ 2^{\frac{1+\varepsilon}{\varepsilon}}\alpha\triangle_{j} ^{ \frac{1+\varepsilon}{\varepsilon}}T_{j-1}}{K-1}\right)} \le \sqrt{\frac{2^{\frac{1+\varepsilon}{\varepsilon}}\alpha \triangle_j^{\frac{1+\varepsilon}{\varepsilon}}T_{j-1}}{K}} \\
    &\leq\frac{1}{2M},
    \label{eq:lower-lem4}
\end{align}
where we define $\tau_{k}$ to be the number of pulls of arm $k$ before time $T_{j-1}$. In above inequality, (a) is due to similar argument in Eq.(\ref{eq:static-1}), (b) is due to Jensen's inequality, and $\sum_{k=1}^{K-1}\tau_{k}\leq{}T_{j-1}$ holds almost surely. The previous two inequalities imply that
\begin{align*}
    |P_{M}(A_{j})-p_{j}|\leq{}\frac{1}{K-1}\sum_{k=1}^{K-1}|P_{M}(A_{j})-P_{j,k}(A_{j})|\leq{}\frac{1}{2M}.
\end{align*}
\end{proof}

\subsection{Proof of Lemma \ref{lem:lower-mini}}
\begin{proof}

 Without loss of generality, we assume that $j \in [M-1] $; while the case $j = M $ is analogous. For any $k \in [K-1]$, consider the following family of reward distributions $\mathcal{P}_{j,k} = (Q_{j,k,\ell})_{\ell \in [K]}$ ,where we define $ Q_{j,k,k} = P_{j,k}$, and for $ \ell \neq k $, let $ Q_{j,k,\ell}$ be the modification of $ P_{j,k}$ where $ 3\triangle_{j} $ is added to the mean of the $\ell $-th component of $ P_{j,k} $.Then, we have the following observations:
\begin{enumerate}
    \item For each $ \ell \in [K] $, arm $ \ell $ is the optimal arm under reward distribution $Q_{j,k,\ell}$;
    \item For each $ \ell \in [K] $, pulling an arm $ \ell' \neq \ell $ incurs a regret at least $ \triangle_{j}$ under reward distribution $ Q_{j,k,\ell} $;
    \item For each $\ell \neq k $, the distributions $Q_{j,k,\ell}$ and $ Q_{j,k,k}$ only differ in the $\ell$-th component.
\end{enumerate}

By our first two observations, the similar arguments in Eq.(\ref{eq:static-1}) yield to
\begin{align*}
\sup_{\{\mu_i\}_{i=1}^{K} : \triangle_{i} \leq K^{\frac{\varepsilon}{1+\varepsilon}}} \mathbb{E}[R_{T}(\pi)] \geq \triangle_{j} \sum_{t=1}^{T} \frac{1}{K} \sum_{\ell=1}^{K} Q_{j,k,\ell}^{t}(\pi_{t} \neq \ell),
\end{align*}

where $ Q_{j,k,\ell}^{t}$ denotes the distribution of observations available at time $t$ under reward distribution $Q_{j,k,\ell}$, and $\pi_{t}$ denotes the policy at time $t$. And we could lower the bound above as
\begin{align}
\sup_{\{\mu_i \}_{i=1}^{K} : \triangle_{i} \leq K^{\frac{\varepsilon}{1+\varepsilon}}} \mathbb{E}[R_{T}(\pi)] & \overset{(a)}{\geq} \triangle_{j} \sum_{t=1}^{T} \frac{1}{K} \sum_{\ell \neq k} \int \min\{dQ_{j,k,k}^{t}, dQ_{j,k,\ell}^{t}\} \notag \\
& \geq \triangle_{j} \sum_{t=1}^{T_{j}} \frac{1}{K} \sum_{\ell \neq k} \int \min\{dQ_{j,k,k}^{t}, dQ_{j,k,\ell}^{t}\} \notag \\
& \overset{(b)}{\geq} \triangle_{j} T_{j} \cdot \frac{1}{K} \sum_{\ell \neq k} \int \min\{dQ_{j,k,k}^{T_{j}}, dQ_{j,k,\ell}^{T_{j}}\} \notag \\
& \geq \triangle_{j} T_{j} \cdot \frac{1}{K} \sum_{\ell \neq k} \int_{A_{j}} \min\{dQ_{j,k,k}^{T_{j}}, dQ_{j,k,\ell}^{T_{j}}\} \notag \\
& \overset{(c)}{=} \triangle_{j} T_{j} \cdot \frac{1}{K} \sum_{\ell \neq k} \int_{A_{j}} \min\{dQ_{j,k,k}^{T_{j-1}}, dQ_{j,k,\ell}^{T_{j-1}}\}, \label{eq:18}
\end{align}
where (a) is due to the proof of Lemma 3 in \cite{gao2019batched}  and considering a star graph on $[K]$ with center $k$, and (b) is due to the identity $\int \min\{dP, dQ\} = 1 - \operatorname{TV}(P, Q)$ and the data processing inequality of the total variation distance, and for step (c) we note that when $A_j = \{ t_{j-1} < T_{j-1}, t_j \geq T_j \}$ holds, the observations seen at time $T_j$ are the same as those seen at time $T_{j-1}$. To lower bound the final quantity, we further have
\begin{align}
\int_{A_j} \min \{ dQ^{T_{j-1}}_{j,k,k}, dQ^{T_{j-1}}_{j,k,\ell} \} & = \int_{A_j} \frac{dQ^{T_{j-1}}_{j,k,k} + dQ^{T_{j-1}}_{j,k,\ell} - |dQ^{T_{j-1}}_{j,k,k} - dQ^{T_{j-1}}_{j,k,\ell}|}{2} \notag \\
& = \frac{Q^{T_{j-1}}_{j,k,k}(A_j) + Q^{T_{j-1}}_{j,k,\ell}(A_j)}{2} - \frac{1}{2} \int_{A_j} |dQ^{T_{j-1}}_{j,k,k} - dQ^{T_{j-1}}_{j,k,\ell}| \notag \\
& 
\overset{(d)}{\geq} \left( Q^{T_{j-1}}_{j,k,k}(A_j) - \frac{1}{2} TV(Q^{T_{j-1}}_{j,k,k}, Q^{T_{j-1}}_{j,k,\ell}) \right) - TV(Q^{T_{j-1}}_{j,k,k}, Q^{T_{j-1}}_{j,k,\ell}) \notag \\
& 
\overset{(e)}{=} P_{j,k}(A_j) - \frac{3}{2} TV(Q^{T_{j-1}}_{j,k,k}, Q^{T_{j-1}}_{j,k,\ell}),\label{eq:19}
\end{align} 
where (d) follows from $|P(A) - Q(A)| \leq TV(P, Q)$, and in (e) we have used the fact that the event $A_j$ can be determined by the observations up to time $T_{j-1}$ . The similar argument in Eq.(\ref{eq:lower-lem4}) yields that,
\begin{align}
\frac{1}{K} \sum_{\ell \neq k} TV(Q^{T_{j-1}}_{j,k,k}, Q^{T_{j-1}}_{j,k,\ell}) & \leq \frac{1}{K} \sum_{\ell \neq k} \sqrt{1 - \exp(-D_{KL}(Q^{T_{j-1}}_{j,k,k}||Q^{T_{j-1}}_{j,k,\ell}))} \notag \\
&
\leq \frac{1}{K} \sum_{\ell \neq k} \sqrt{1 - \exp( -6\triangle_{j}\gamma \mathbb{E}_{P_{j,k}[\tau_{\ell}]} )}\notag \\
&
\leq \frac{1}{K} \sum_{\ell \neq k} \sqrt{1 - \exp( -\alpha\triangle_{j}^{\frac{1+\varepsilon}{\varepsilon}}\mathbb{E}_{P_{j,k}[\tau_{\ell}]} )}\notag \\
&
\leq \frac{K-1}{K} \sqrt{1 - \exp\left(\frac{ -\alpha\triangle_{j}^{\frac{1+\varepsilon}{\varepsilon}} \sum_{\ell \neq k} \mathbb{E}_{P_{j,k}}[\tau_{\ell}]}{K-1} \right)} \notag \\
&
\leq \frac{K-1}{K} \sqrt{1 - \exp\left(\frac{ -\alpha\triangle_{j}^{\frac{1+\varepsilon}{\varepsilon}}T_{j-1}}{K-1}\right)} \leq \sqrt{ \frac{\alpha\triangle_{j}^{\frac{1+\varepsilon}{\varepsilon}}T_{j-1}}{K}} \notag \\
&
\leq \frac{1}{12M},\label{eq:20}
\end{align}
where the third inequality is due to the concavity of $x \mapsto \sqrt{1 - e^{-x}}$ for $x \geq 0$, the fourth inequality follows from $\sum_{\ell \neq k} \tau_{\ell} \leq T_{j-1}$ almost surely.

Combining Eq.(\ref{eq:18}), (\ref{eq:19}) and (\ref{eq:20}), we conclude that
\begin{align*}
\sup_{\{ \mu_i \}_{i=1}^{K}: \triangle_i \leq K^{\frac{\varepsilon}{1+\varepsilon}}} \mathbb{E}[R_T(\pi)] &\geq \triangle_j T_j \left( \frac{P_{j,k}(A)}{2} - \frac{1}{8M} \right) \\
&\geq K^{\frac{\varepsilon}{1+\varepsilon}} T^{\frac{1}{1+\varepsilon-\varepsilon\left( \frac{\varepsilon}{1+\varepsilon} \right)^{M-1}}} 
\cdot \frac{2^{-\frac{3\varepsilon+1}{1+\varepsilon}}}{2 (12\sqrt{\alpha} M)^{\frac{2\varepsilon}{1+\varepsilon}}} \left( \frac{P_{j,k}(A)}{2} - \frac{1}{8M} \right).
\end{align*}

Note that the above inequality holds for any $k \in [K-1]$, averaging over $k \in [K-1]$ yields
\begin{align*}
\sup_{\{ \mu_i \}_{i=1}^{K}: \triangle_i \leq K^{\frac{\varepsilon}{1+\varepsilon}}} \mathbb{E}[R_T(\pi)] & \geq K^{\frac{\varepsilon}{1+\varepsilon}} T^{\frac{1}{1+\varepsilon-\varepsilon\left( \frac{\varepsilon}{1+\varepsilon} \right)^{M-1}}} \cdot \frac{2^{-\frac{3\varepsilon+1}{1+\varepsilon}-1}}{(12\sqrt{\alpha} M)^{\frac{2\varepsilon}{1+\varepsilon}}} \left( \frac{\sum_{k=1}^{K-1}P_{j,k}(A)}{2(K-1)} - \frac{1}{8M} \right) \\
&
\geq \frac{1}{16\cdot 2^{\frac{3\varepsilon+1}{1+\varepsilon}}\cdot (\sqrt{\alpha} M)^{\frac{3\varepsilon}{1+\varepsilon}}} \cdot K^{\frac{\varepsilon}{1+\varepsilon}} T^{\frac{1}{1+\varepsilon-\varepsilon\left( \frac{\varepsilon}{1+\varepsilon} \right)^{M-1}}},
\end{align*}
where in the last inequality we use that $p_j \geq \frac{1}{2M}$. 
The coefficient $\frac{1}{16\cdot 2^{\frac{3\varepsilon+1}{1+\varepsilon}}\cdot (\sqrt{\alpha} )^{\frac{3\varepsilon}{1+\varepsilon}}}$ is larger than some numerical constant $\kappa$ since $\alpha = 6 2^\frac{1}{\varepsilon}$.
Hence, we have proved \textbf{ Lemma \ref{lem:lower-mini}}.
    
\end{proof}

\subsection{Proof of Lemma \ref{lem:upper-lips}}
\begin{proof}
    We first estimate the probability of event $E$. By \textbf{Lemma} \ref{heavy-tail:property} and Jensen's inequality, mean estimators can approximate the true values. For any $m \in [M]$, $C \in A_m$, and $n_m \in \mathbb{N}$, with probability at least $1-2 T^{-3}$, we have
    \begin{align*}
        | \mathbb{E}_C [\mu(y)] - \hat{\mu}_C(n_m) |
        \le
        v^{\frac{1}{1+\varepsilon}} \left( \frac{c \log(T^3)}{n_m} \right)^\frac{\varepsilon}{1+\varepsilon},
    \end{align*}
    where the expectation $\mathbb{E}_C$ means $y$ chooses the center of set $C$. Here we choose $n_m = 3 c v^\frac{1}{\varepsilon} \log T r_m^{-\frac{1+\varepsilon}{\varepsilon}}$, the above inequality is
    \begin{align*}
        \left| \mathbb{E}_C [\mu(y)] - \hat{\mu}_C(n_m) \right|
        \le r_m.
    \end{align*}
    According to 1-Lipschitz, for any $x \in C$ at batch $m$, we have $ | \mu(x) - \mathbb{E}_C [\mu(y)] | \le r_m $.
    That is to say, for $\forall C \in A_m$, with probability at least $1-2 T^{-3}$,
    \begin{align*}
        \left|\mu(x)-\hat{\mu}_C(n_m)\right| \le 2 r_m, \quad \forall x \in C.
    \end{align*}

    Event $E$ states that the above bound holds true for all $m$ and $C$. We can count the number of all set $C$ in $A_1, \cdots, A_{M-1}$, and find that it is no larger than $T$ since each $C$ is sampled at least once. Taking a union bound gives
    \begin{align*}
        \mathbb{P} (E) \le 1 - 2 T^{-2}.
    \end{align*}

    Then we proof that under event $E$, the best arm $\star$ will not be eliminated. At batch $m \le M-1$, we use $C_{\star}$ to denote the cube containing $\star$ in $A_m$. For any $C$ in $A_m$ and $x \in C$,
    \begin{align*}
        \hat{\mu}_C(n_m) - \hat{\mu}_{C_\star}(n_m) 
        \le
        \mu(x) + 2 r_m - (\mu_\star -2r_m)
        \le 4 r_m.
    \end{align*}
    Finally, $\mathbb{P}(\mathcal{E}) = \mathbb{P}(B \cup E)=\mathbb{P}(E) \le 1-2T^{-2}$ finishes the proof.
\end{proof}


\end{document}